\documentclass{article}

\PassOptionsToPackage{numbers, compress}{natbib}
\usepackage[preprint]{neurips_2020}

\usepackage[utf8]{inputenc} % allow utf-8 input
\usepackage[T1]{fontenc}    % use 8-bit T1 fonts
\usepackage{hyperref}       % hyperlinks
\usepackage{url}            % simple URL typesetting
\usepackage{booktabs}       % professional-quality tables
\usepackage{amsfonts}       % blackboard math symbols
\usepackage{nicefrac}       % compact symbols for 1/2, etc.
\usepackage{microtype}      % microtypography

\usepackage{amsmath}
\usepackage{subcaption}
\usepackage{dsfont}
\usepackage{enumitem}
\usepackage{graphicx}
\usepackage{multirow}
\usepackage{comment}
\usepackage{amsthm}
\usepackage{makecell}
\usepackage{bm}
\usepackage[ruled,vlined,linesnumbered]{algorithm2e}
\usepackage{algorithmic}
\usepackage{threeparttable}
\usepackage{adjustbox}

\usepackage{wrapfig}
\newtheorem{theorem}{Theorem}
\newtheorem{corollary}{Corollary}
\newtheorem{proposition}{Proposition}
\newtheorem*{remark}{Remark}

\SetKwInput{KwRequire}{Require}

\title{Adversarial Likelihood-Free Inference\\on Black-Box Generator}

\author{%
	Dongjun~Kim, Weonyoung~Joo, Seungjae~Shin, Kyungwoo~Song, Il-Chul~Moon\\
  KAIST, Republic of Korea\\
  \texttt{\{dongjoun57, es345, tmdwo0910, gtshs2, icmoon\}@kaist.ac.kr}
}

\begin{document}
	
	\maketitle
	
	\begin{abstract}
		Generative Adversarial Network (GAN) can be viewed as an implicit estimator of a data distribution, and this perspective motivates using the adversarial concept in the true input parameter estimation of black-box generators. While previous works on \textit{likelihood-free inference} introduces an implicit proposal distribution on the generator input, this paper analyzes theoretic limitations of the proposal distribution approach. On top of that, we introduce a new algorithm, Adversarial Likelihood-Free Inference (ALFI), to mitigate the analyzed limitations, so ALFI is able to find the posterior distribution on the input parameter for black-box generative models. We experimented ALFI with diverse simulation models as well as pre-trained statistical models, and we identified that ALFI achieves the best parameter estimation accuracy with a limited simulation budget.
	\end{abstract}
	
	\section{Introduction} \label{sec:Introduction}
	Generative Adversarial Network (GAN) is highlighted recently for its success on the implicit estimation of the data distribution. In GAN, the generator is jointly learned with the discriminator, so the generator becomes a trainable and fine-tunable model. In contrast to training both the generator and the discriminator in GAN, there has been a line of work on applying the adversarial framework on the pre-trained and fixed generator to estimate the optimal input of the generator \cite{louppe2019adversarial}. For example, a simulation model can be considered as a generator that is not successfully integrated into the adversarial concept. Researchers are interested in inferring the posterior distribution of a simulation input parameter with a snapshot of a validation observation from the real-world \cite{meeds2015optimization, gutmann2016bayesian, louppe2019adversarial}.
	
	\begin{wrapfigure}{r}{0.3\textwidth}
		\vskip -0.2in
		\centering
		\includegraphics[width=\linewidth]{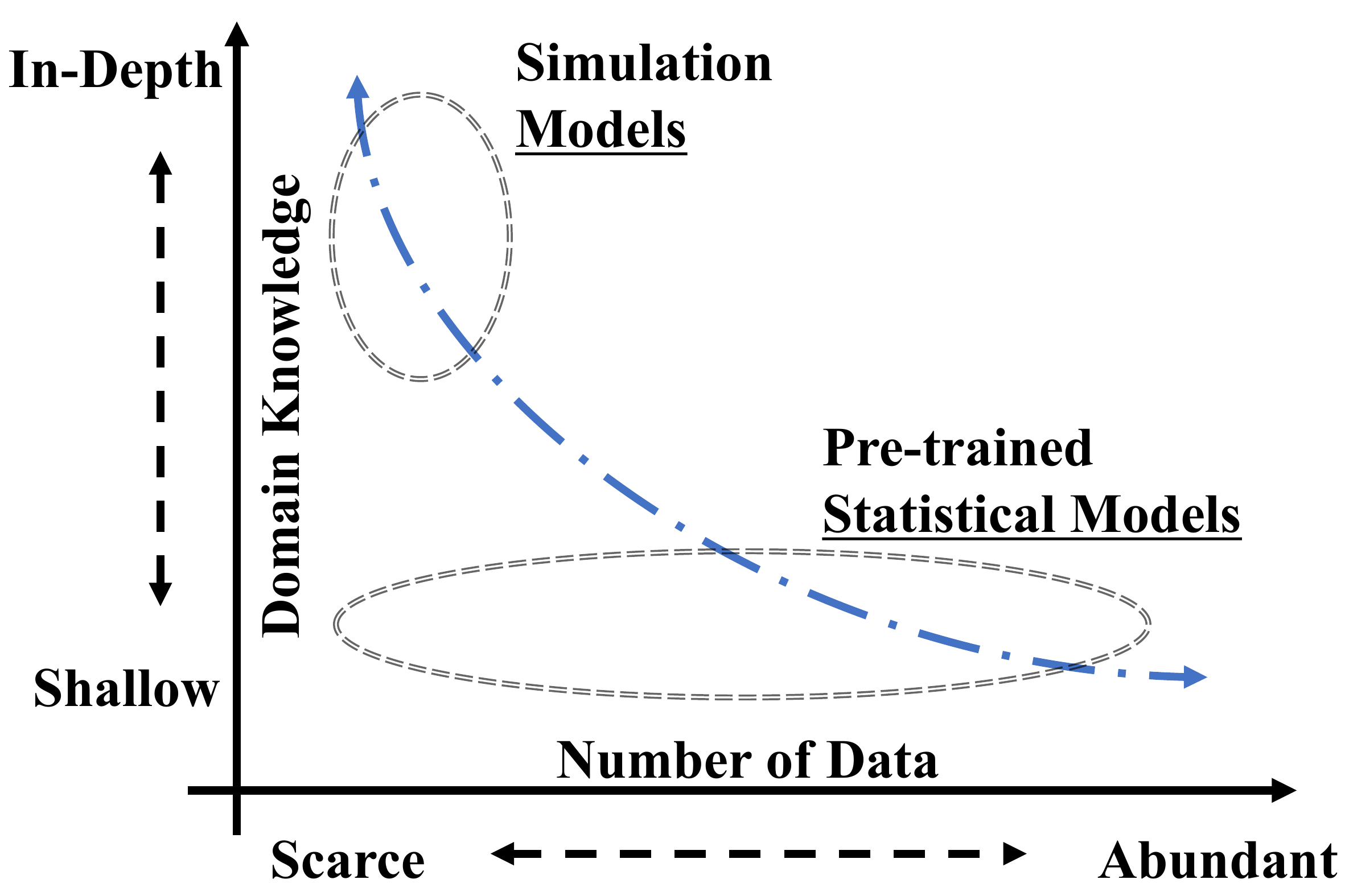}
		\caption{Black-box generative models}
		\label{img:DomainKnowledge}
		\vskip -0.1in
	\end{wrapfigure}
	
	Before we move on, we define a \textit{black-box generative model} to present our interested generator type clearly. The \textit{black-box generative model} ($g$) indicates a generative model that has three properties. First, the model's internal structure is designed \textit{before} the inference stage by domain experts or as another statistical model. Second, the internal coefficients ($\omega$) do not change since the coefficients are obtained by the domain-specific knowledge or through a separate learning process. Third, the internal process contains the inherent stochasticity ($u$) that forms various sample paths for each of the model execution. For example, continuous, discrete, and agent-based simulation models can be black-box generators that produce the stochastic trajectories of modeled states from simulations. As another example, a pre-trained and fixed de-convolutional neural network can be another practical case of the black-box generators.
	
	The research on the black-box generator emphasizes the regeneration of the observation, which is formulated as the inference on the posterior distribution $p(\theta|x_{obs})$, where $\theta\in\Theta\subseteq\mathbb{R}^{d}$ is a $d$-dimensional generator input parameter on a compact space $\Theta$, and where $x_{obs}\in\mathbb{P}_{r}$ is the \textit{single} instance of the real-world data to reconstruct from the real-world data distribution $\mathbb{P}_{r}$. % is the real-world data distribution. 
The likelihood $p(x_{obs}|\theta)$ of the black-box generator requires the integration over the aforementioned generation stochasticity, or a nuisance variable $u\in\mathbb{R}^{m}$, because $u$ determines the sample path of the generator. However, $u$ is unknown in general, and the integration over $u$ is likely to be intractable. Therefore, the Bayesian inference under black-box generators mainly focuses on estimating the intractable likelihood, called \textit{likelihood-free inference}. This paper provides theoretic analysis on previous research, and suggests a new algorithm of \textit{likelihood-free inference} under the adversarial setting.

%	The contribution of this paper is four-fold.
%	\begin{enumerate}
%		\item This paper analyzes the \textit{gradient vanishing problem} and the \textit{implicit relation problem} of the previous research on \textit{likelihood-free inference}.
%		\item This paper provides a formula of the intractable likelihood as a $1$-dimensional density of a random variable from the discriminator.
%		\item This paper suggests a new \textit{likelihood-free inference} that uses the adversarial framework.
%		\item This paper proves the convergence of the limiting distribution to the posterior distribution of a Markov chain with a dynamically updated transition kernel under the Metropolis-Hastings algorithm.
%	\end{enumerate}
	
	\section{Previous Research} \label{sec:PreviousResearch}
	
	In \textit{likelihood-free inference} community, the summary statistics $s:\mathbb{R}^{p}\rightarrow\mathbb{R}^{q}$ extracts a set of statistics from either observation $x_{obs}$ or generated fake data $g(\theta,u|\omega)$, and the discrepancy function $d:\mathbb{R}^{q}\times\mathbb{R}^{q}\rightarrow\mathbb{R}$ measures how $s\big(g(\theta,u|\omega)\big)$ deviates from $s(x_{obs})$. Throughout the paper, we assume that the function $s$ extracts the \textit{sufficient} statistics \cite{beaumont2010approximate, casella2002statistical, kolmogorov1942determination} that assure the identity of the posteriors under the summary statistics as $p(\theta|x_{obs})=p\big(\theta|s(x_{obs})\big)$. In addition, we assume the summary statistics to be embedded in the generator, so the generated data $g(\theta,u|\omega)$ or the observation $x_{obs}$ becomes the extracted $q$-dimensional summary statistics of the generated raw data or the observed raw data, respectively. The likelihood, $p(x_{obs}|\theta)=\int p(x_{obs}|u,\theta)p_{U}(u)du=\int \delta\big(d(g(\theta,u|\omega),x_{obs})\big)p_{U}(u)du$, is intractable since the distribution $p_{U}(u)$ and the level set, $\big\{u\mid d\big(g(\theta,u|\omega),x_{obs}\big)=0\big\}$, are generally not known in a black-box generator.
	
	\textbf{Approximate Bayesian Computation:} Approximate Bayesian Computation (ABC) \cite{tavare1997inferring, fu1997estimating, marjoram2003markov, sisson2007sequential, beaumont2009adaptive} estimates the likelihood through Monte-Carlo methods by approximating the singular distribution $\delta$ with mollifier kernels $\{K_{\epsilon}\}_{\epsilon>0}$, so that $p(x_{obs}|\theta)=\lim_{\epsilon\downarrow 0}\mathbb{E}_{u\sim p_{U}}\big[K_{\epsilon}(u;\theta)\big]$. A case of the ABC algorithm is the Rejection ABC \cite{tavare1997inferring} that uses the boxcar kernel \cite{stein2011functional}: $p(x_{obs}|\theta)=\lim_{\epsilon\downarrow 0}\frac{1}{\vert B_{\epsilon}(x_{obs})\vert}\mathbb{E}_{u}\big[1_{B_{\epsilon}(x_{obs})}\big(g(\theta,u|\omega)\big)\big]$, where $B_{\epsilon}(x_{obs})$ is the $\epsilon$-ball $\{x:d(x,x_{obs})<\epsilon\}$.
	
	\textbf{Bayesian Optimization Likelihood-Free Inference:} Bayesian Optimization Likelihood-Free Inference (BOLFI) \cite{gutmann2016bayesian} infers the predictive distribution of the discrepancy $d\big(g(\theta,u|\omega),x_{obs}\big)$, using a Gaussian process regression with a dataset $\mathcal{D}_{1:t}=\big\{(\theta_{i},d(g(\theta_{i},u_{i}|\omega),x_{obs})\big\}_{i=1}^{t}$, and BOLFI samples a new parameter $\theta_{t+1}$ by the Bayesian optimization. The likelihood of the discrepancy being less than a threshold $\epsilon$ is proportional to $p(x_{obs}|\theta)\propto\Phi(\frac{\epsilon-\hat{\mu}(\theta)}{\hat{\sigma}(\theta)})$, where $\Phi$ is the cumulative distribution function of the standard Gaussian distribution, and where $\hat{\mu}$ and $\hat{\sigma}$ are the mean and the standard deviation of the predictive distribution of the Gaussian process regression.
	
	\section{Preliminary: Problems of Implicit Proposal Distribution}\label{sec:Preliminary}
	
	\subsection{Motivation of Implicit Proposal Distribution}
	
\begin{figure}[t]
		\begin{center}
			\begin{subfigure}{.24\linewidth}
				\vskip 0.09in
				\centering
				\includegraphics[width=0.92\linewidth]{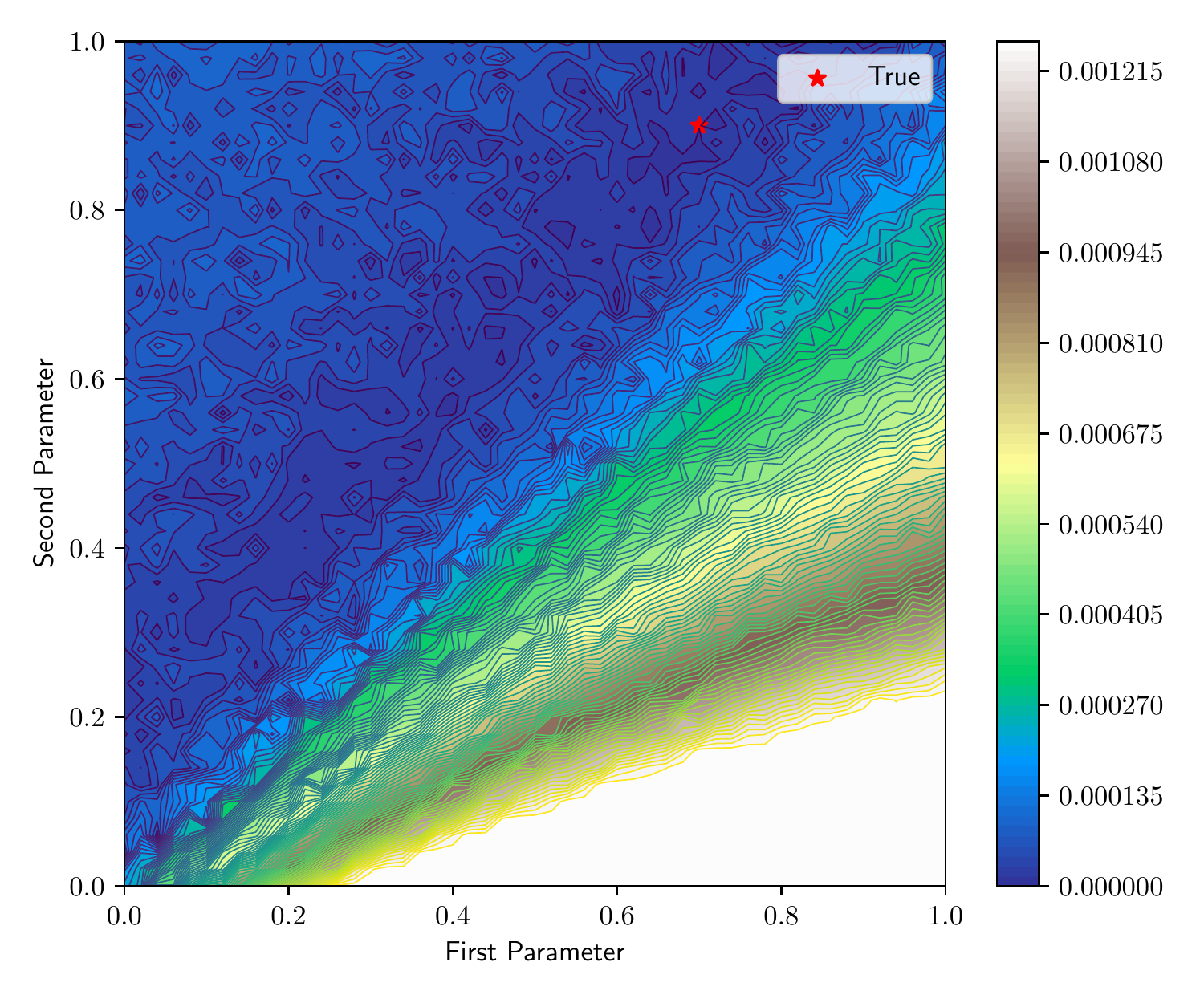}
				\subcaption{Discrepancy Map}
			\end{subfigure}
			\begin{subfigure}{.24\linewidth}
				\centering
				\includegraphics[width=\linewidth]{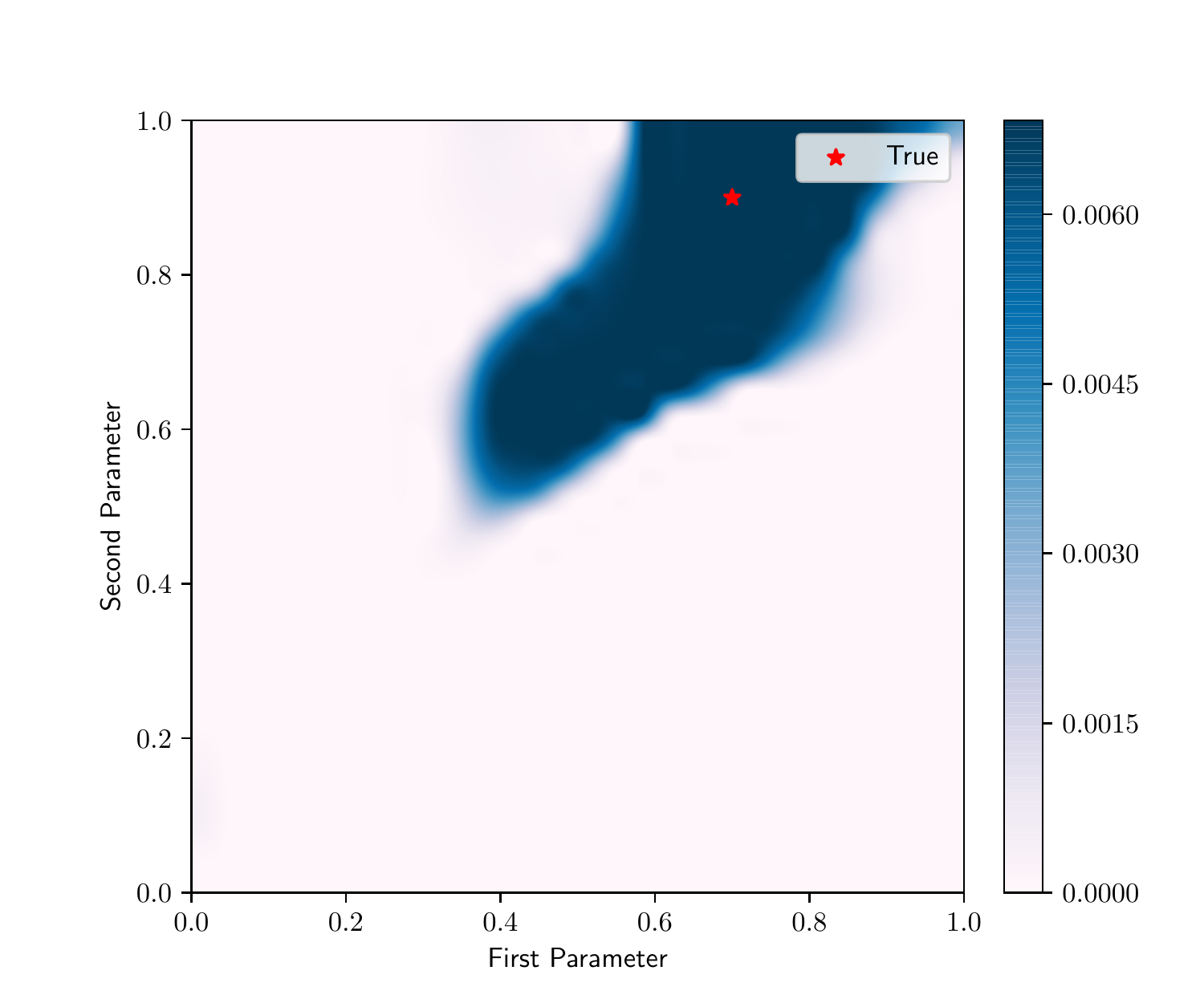}
				\subcaption{BOLFI Posterior}
			\end{subfigure}
			\begin{subfigure}{.24\linewidth}
				\centering
				\includegraphics[width=\linewidth]{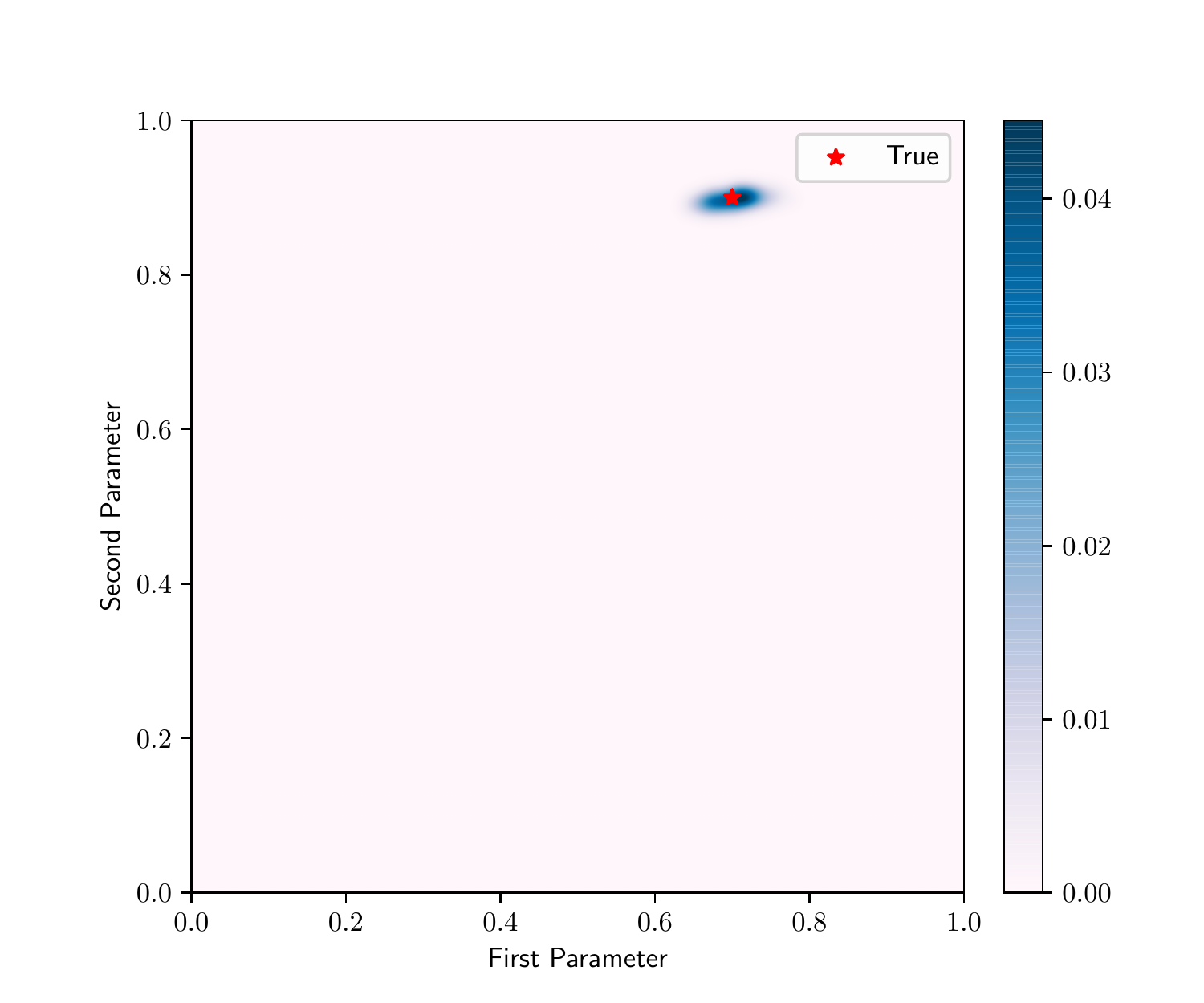}
				\subcaption{ALFI Posterior}
			\end{subfigure}
			\begin{subfigure}{.24\linewidth}
				\vskip 0.1in
				\centering
				\includegraphics[width=\linewidth]{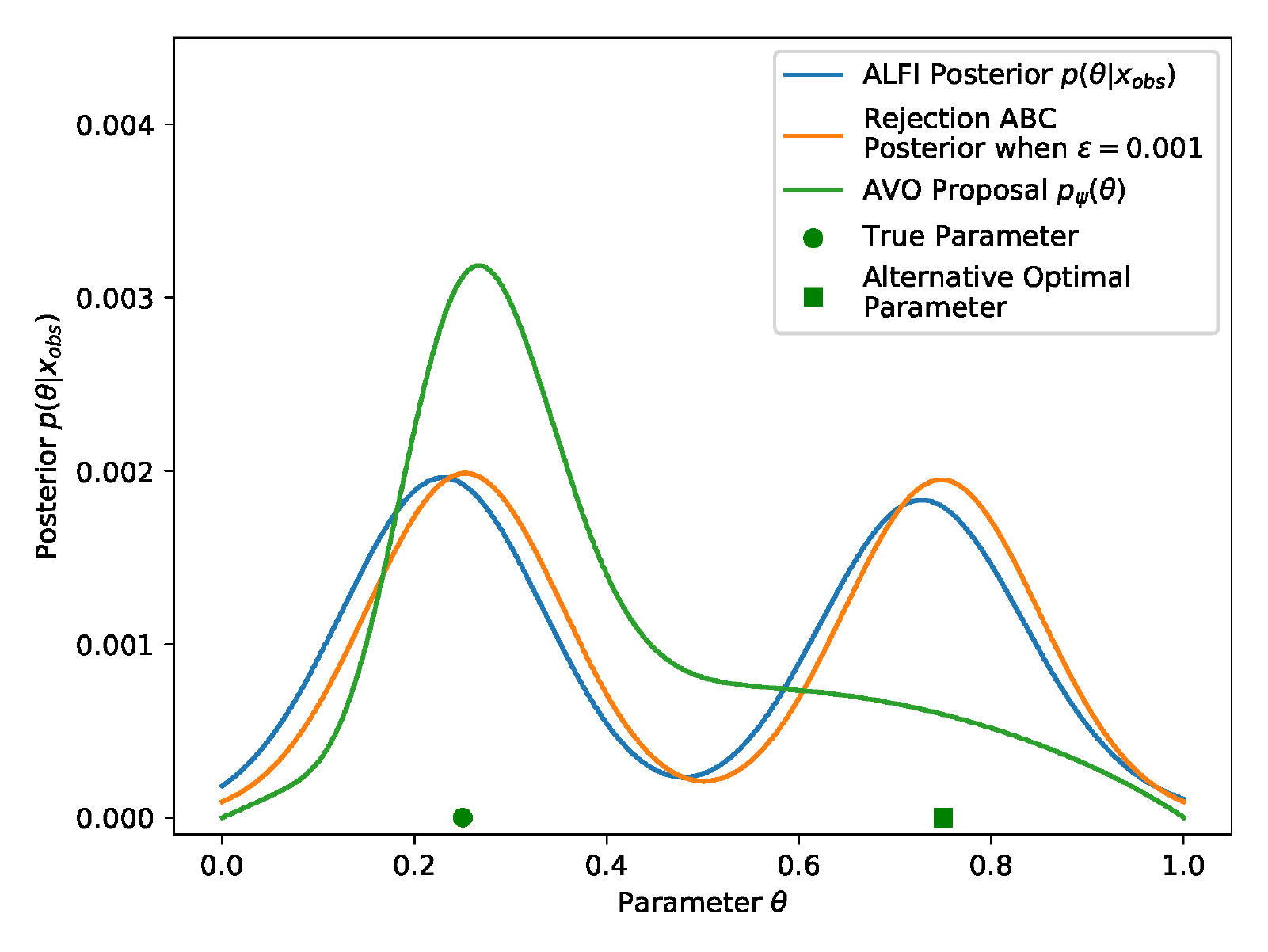}
				\subcaption{AVO Posterior}
			\end{subfigure}
			\caption{(a-c) The Susceptible-Infectious-Recovered simulation model result. (a) The value at $(\theta_{1},\theta_{2})$ represents the discrepancy $\Vert x_{obs}-g((\theta_{1},\theta_{2}),u)\Vert_{2}$. (b) BOLFI finds a huge area as the candidate region for the true parameter $\theta^{*}$. (c) ALFI captures the true parameter $\theta^{*}$ within a tiny region. (d) The inferred posterior with simulation model $g(\theta,u)=(\theta-0.5)^{2}+u$, where $u\sim\mathcal{N}(0,10^{-4})$.}
			\label{img:Multi-modal}
		\end{center}
		\vskip -0.2in
	\end{figure}

	Figure \ref{img:Multi-modal} (a) illustrates the discrepancy ($d$) landscape of the Susceptible-Infectious-Recovered (SIR) simulation model \cite{diekmann2000mathematical}, where the discrepancy is defined as the Euclidean measure. The choice of the summary statistics and the discrepancy measure are crucial in \textit{likelihood-free inference}, since Figure \ref{img:Multi-modal} (b) illustrates the failure of inferring the posterior distribution if the discrepancy landscape is highly rugged, and if the landscape has a plateau near the true parameter $\theta^{*}$.
	
	This failure leads \textit{likelihood-free inference} community to investigate the adaptive selection of summary statistics and a discrepancy measure. The \textit{likelihood-free inference} under adversarial setting partially solves the selection problem by constructing discrepancy measure as a discriminator network. Besides, in some black-box generators, it is possible to put raw data into the discriminator network, without extracting the summary statistics. However, \textit{likelihood-free inference} under the adversarial framework is rarely proposed since the gradient with respect to the input parameter $\theta$ is not backpropagated through an implicitly defined generator that has no closed-form solution. Recently, \citet{louppe2019adversarial} overcomes the backpropagation issue by suggesting the Adversarial Variational Optimization (AVO).
	
	\subsection{Adversarial Variational Optimization}

\begin{figure*}[t]
		\centering
		\begin{subfigure}{.3\linewidth}
			\centering
			\includegraphics[width=\linewidth]{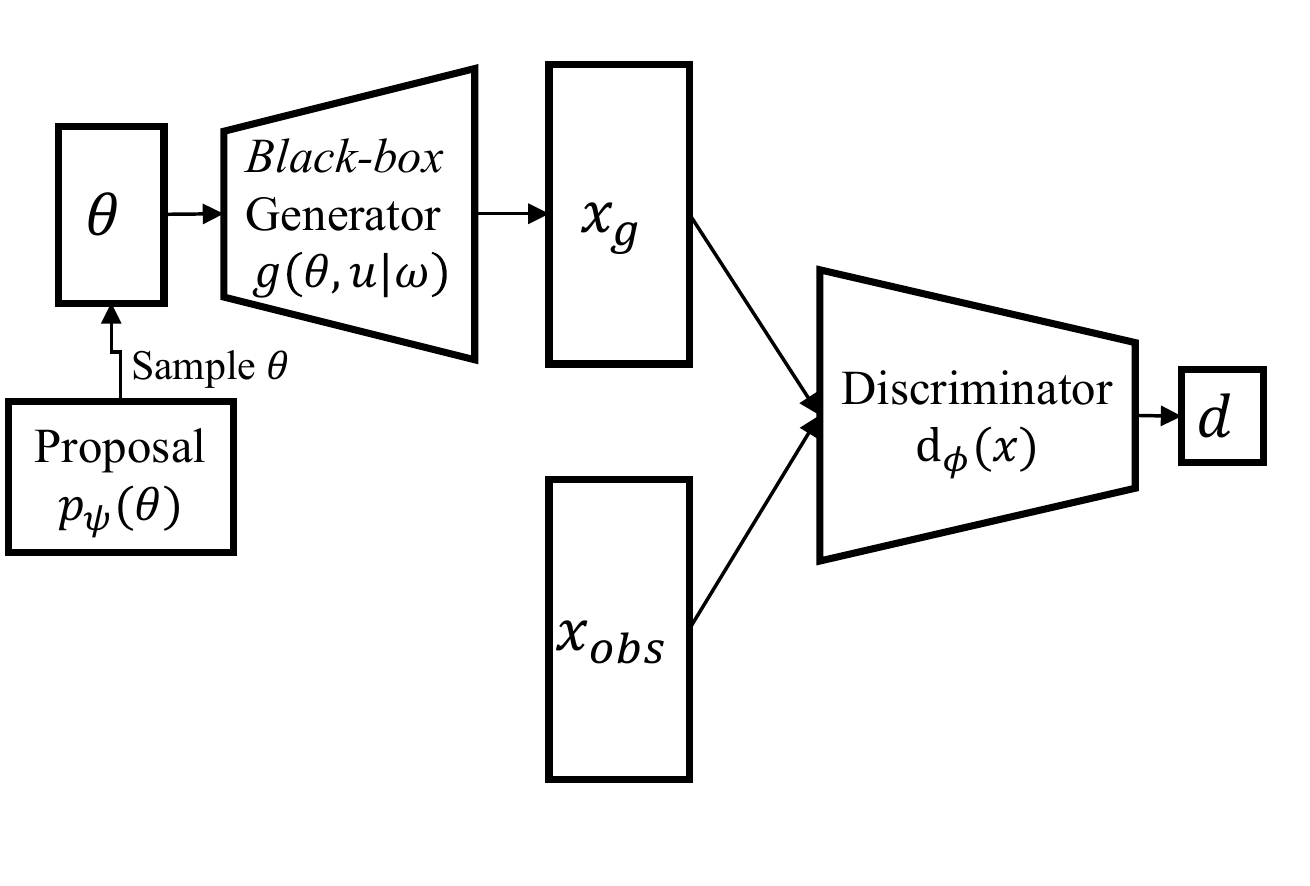}
			\subcaption{AVO Network Structure}
		\end{subfigure}
		\begin{subfigure}{.3\linewidth}
			\centering
			\includegraphics[width=\linewidth]{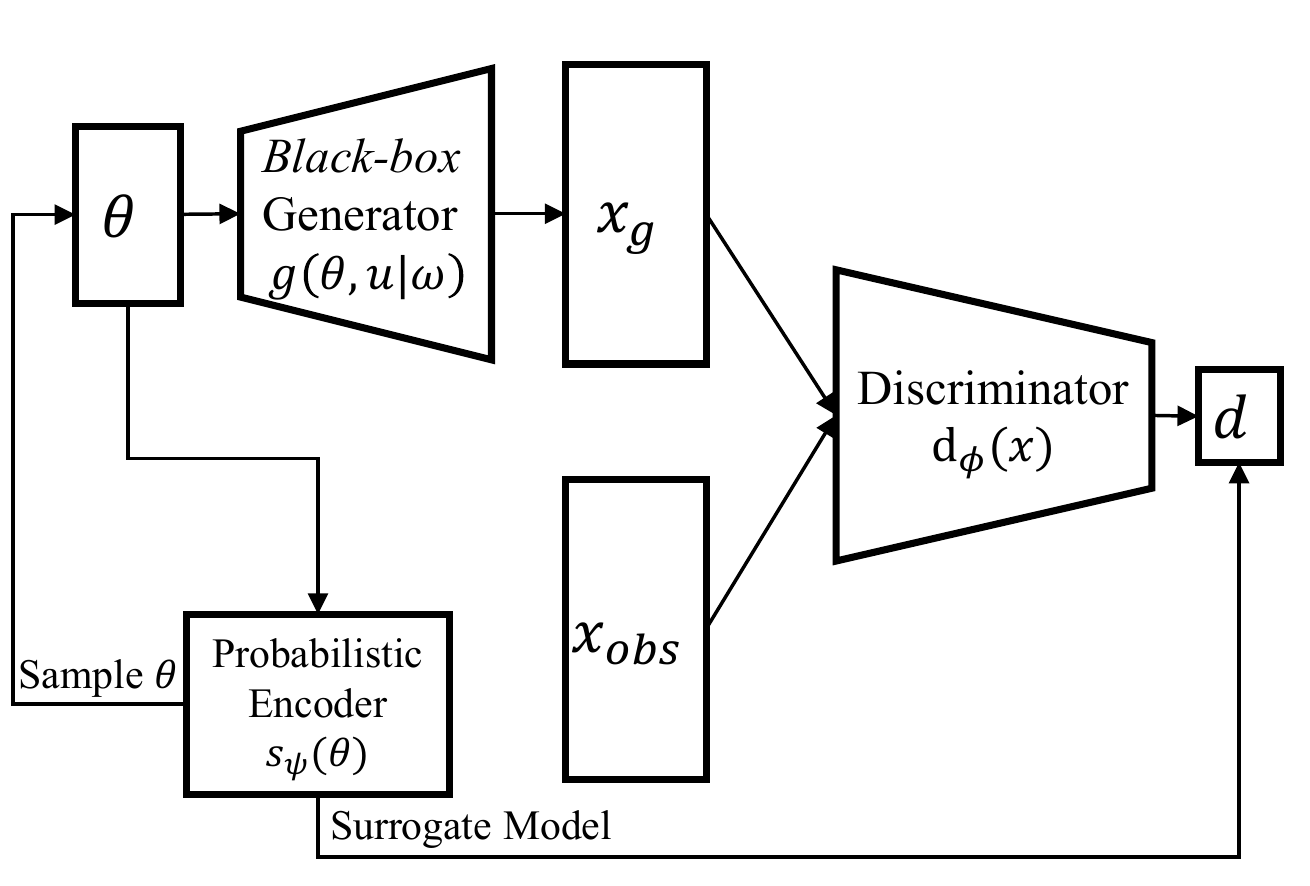}
			\subcaption{ALFI Network Structure}
		\end{subfigure}
		\begin{subfigure}{.19\linewidth}
			\centering
			\includegraphics[width=0.7\linewidth]{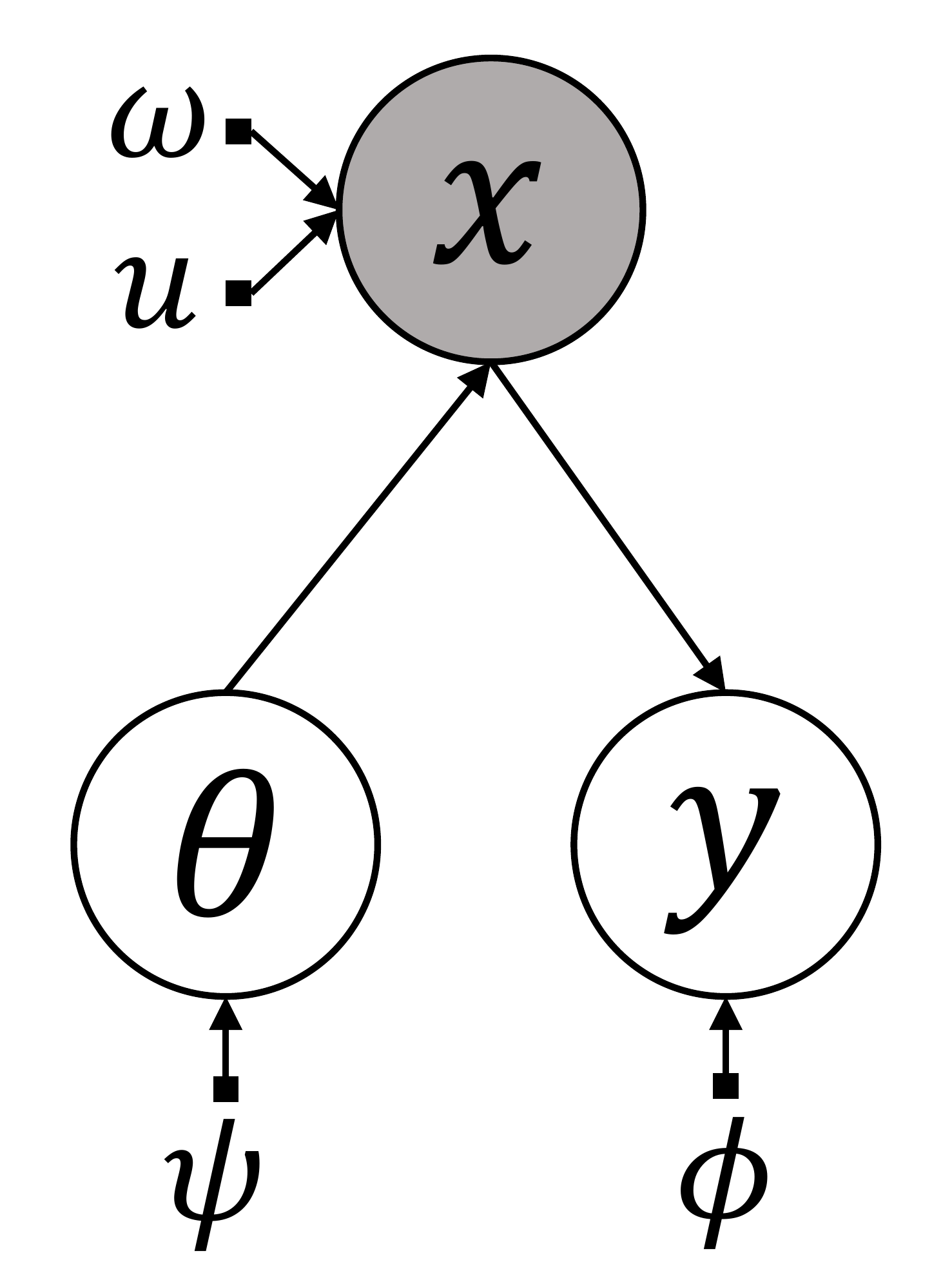}
			\subcaption{AVO Graphical Model}
		\end{subfigure}
		\begin{subfigure}{.19\linewidth}
			\centering
			\includegraphics[width=0.7\linewidth]{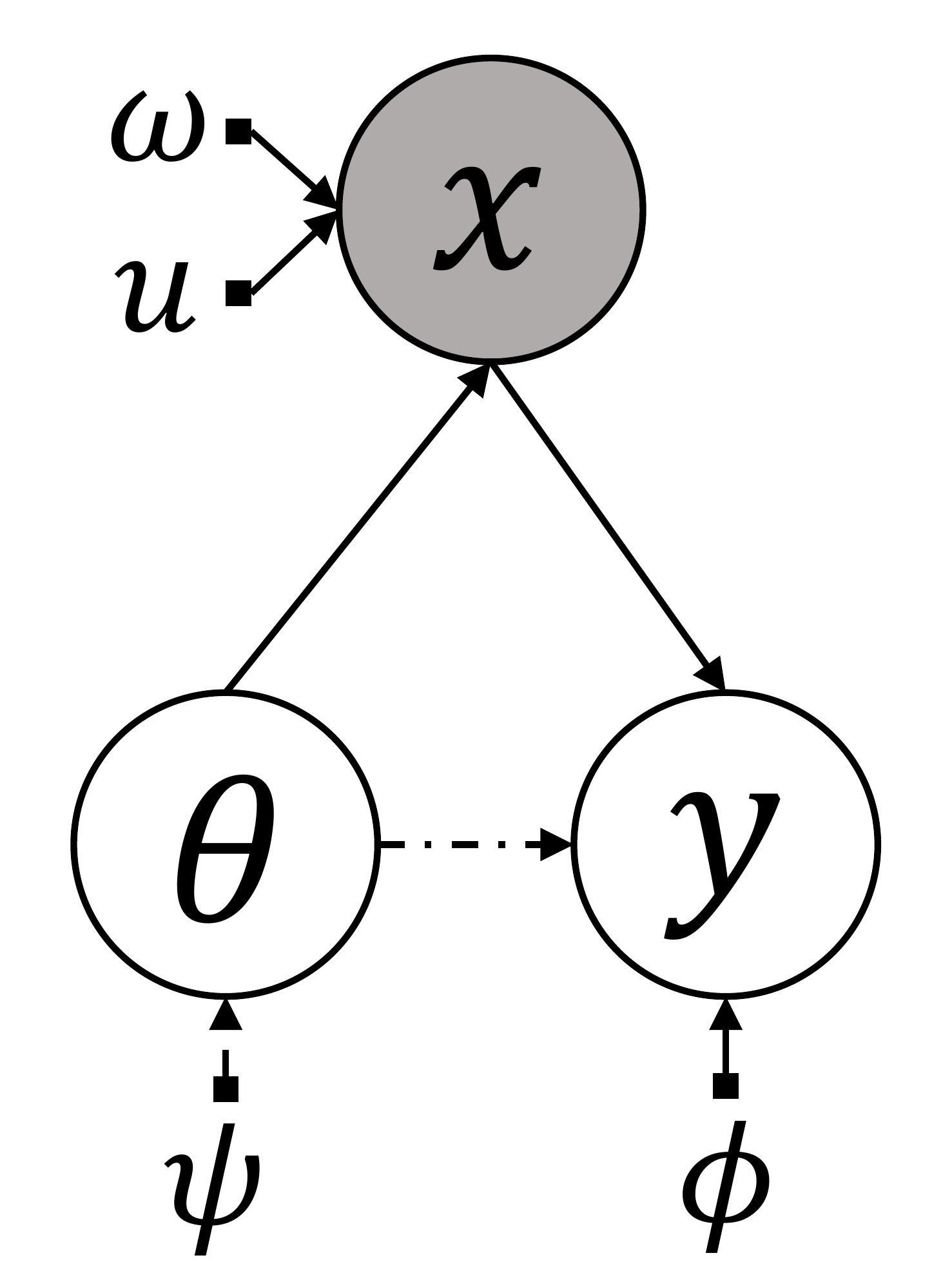}
			\subcaption{ALFI Graphical Model}
		\end{subfigure}
		\caption{AVO and ALFI comparison}
		\label{img:Model_Structures}
		\vskip -0.2in
	\end{figure*}

	The AVO in Figure \ref{img:Model_Structures} (a,c) introduces the implicit proposal distribution $p_{\psi}(\theta)$ for the black-box model input parameter, which enables the backpropagation through a non-differentiable black-box generator by switching the optimization target variable from $\theta$ to $\psi$. AVO has a fixed model internal coefficients $\omega$, so the generator distribution $\mathbb{P}_{g}$ can only be adjusted by inferring the input distribution $p_{\psi}(\theta)$ in order to approximate the data distribution $\mathbb{P}_{r}$.
	
	The minimax function of AVO is given by $V(\psi,\phi)=\mathbb{E}_{x\sim\mathbb{P}_{r}}\big[\log{d_{\phi}(x)}\big]+\mathbb{E}_{\tilde{x}\sim\mathbb{P}_{g}}\big[\log (1-d_{\phi}(\tilde{x}))\big]$, where a fake sample $\tilde{x}$ from the generator distribution $\mathbb{P}_{g}$ is the output of a black-box generator $g(\theta,u|\omega)$, with a sampled input $\theta\sim p_{\psi}$ and a sampled nuisance variable $u\sim p_{U}$. It should be noted that a discriminator $d_\phi(x)$ guides the proposal distribution to enforce the approximation of $\mathbb{P}_{g}$ toward $\mathbb{P}_{r}$. The backpropagation through the generator is calculated by the REINFORCE algorithm, $\nabla_{\psi}V=\mathbb{E}_{\theta\sim p_{\psi}}\big[\nabla_{\psi}p_{\psi}(\theta)\mathbb{E}_{u\sim p_{U}}\big[\log(1-d_{\phi}(g(\theta,u|\omega)))\big]\big]$.
	
	\subsection{Gradient Vanishing Problem of Proposal Distribution Approach}
	\label{sec:GradientVanishing}
	
	The implicit proposal approach in AVO has two significant drawbacks. The first problem is the \textit{gradient vanishing problem} \cite{arjovsky2017principled}. Our analysis is different from the previous theory \cite{arjovsky2017principled} on the gradient \textit{with respect to $\omega$, the internal parameters} of the black-box generator; since we analyze the gradient \textit{with respect to $\psi$, the proposal parameters} of the black-box generator, see Appendix A.
	\begin{proposition}[\emph{Gradient Zero On Optimal Discriminator}]\label{prop:1}
		Let $a:[0,1]\rightarrow\mathbb{R}\cup\{-\infty,\infty\}$ and $b:[0,1]\rightarrow\mathbb{R}\cup\{-\infty,\infty\}$ be strictly increasing and strictly decreasing functions, respectively. Assume the empirical value function is $V_{emp}(\psi,\phi)=\mathbb{E}_{x\in\mathbb{P}_{emp}}[a(d_{\phi}(x))]+\mathbb{E}_{\tilde{x}\in\mathbb{P}_{g}}[b(d_{\phi}(\tilde{x}))]$, where $\mathbb{P}_{emp}$ is the empirical real-world data distribution with a single observation $x_{obs}$.\\
		(i) If $a$ and $b$ are upper bounded, the gradient $\nabla_{\psi}V_{emp}$ of the empirical value function $V_{emp}$ with respect to $\psi$ is always zero under the optimal discriminator.\\
		(ii) If neither $a$ nor $b$ is upper bounded, the empirical value function attains infinity under the optimal discriminator.
	\end{proposition}
	
	\begin{remark}
		If $a(t)=\log{t}$ and $b(t)=\log{(1-t)}$, the value function becomes the vanilla GAN \cite{goodfellow2014generative} loss $V(\psi,\phi)=\mathbb{E}_{x\in\mathbb{P}_{r}}\big[\log{d_{\phi}(x)}\big]+\mathbb{E}_{\tilde{x}\in\mathbb{P}_{g}}\big[\log{(1-d_{\phi}(\tilde{x}))}\big]$. If $a(t)=t$ and $b(t)=-t$, the value function becomes the Wasserstein GAN \cite{arjovsky2017wasserstein} loss $V(\psi,\phi)=\mathbb{E}_{x\in\mathbb{P}_{r}}\big[d_{\phi}(x)\big]-\mathbb{E}_{\tilde{x}\in\mathbb{P}_{g}}\big[d_{\phi}(\tilde{x})\big]$.
	\end{remark}
	Proposition \ref{prop:1} analyzes an extreme and special case that the dataset has a single instance $x_{obs}$, and this scenario occurs in the simulation calibration case (i.e. finding optimal input parameter) because the real-world observation with the same context happens only once. In this case, the \textit{true} value function $V$ is never approximated exactly, and the adversarial framework could gain a saddle point of the empirical value function $V_{emp}$, at best, not the \textit{true} value function $V$.
	\begin{proposition}\label{prop:2}
		Let $a$ and $b$ be strictly increasing and strictly decreasing functions, respectively. For any $\psi$, there exists a constant $M$ such that $\Vert\nabla_{\psi}V\Vert_{2}=\Vert\nabla_{\psi}V_{emp}\Vert_{2}\le M\vert\Theta\vert\max(\vert b(\epsilon)\vert,\vert b(0)\vert)$, where $\Vert\cdot\Vert_{2}$ is the Euclidean norm, if the following conditions satisfy.\\
		(i) The proposal distribution, $p_{\psi}$, is differentiable with respect to $\psi$, and the derivative of $\nabla_{\psi}p_{\psi}$ is continuous with respect to $\theta$.\\
		(ii) The space of the black-box model input parameter, $\Theta$, is compact.\\
		(iii) The discriminator $d_{\phi}$ is $\epsilon$-close to the optimal discriminator $d^{*}$ of the empirical value function: $\Vert d_{\phi}-d^{*}\Vert<\epsilon$, where $\Vert d\Vert=\sup_{x}\vert d(x)\vert$.
	\end{proposition}
	\begin{corollary}[\emph{Gradient Vanishing Near Optimal Discriminator}]\label{cor:1}
		Let $a$ and $b$ be strictly increasing and strictly decreasing functions. Assume that 1) the conditions of Proposition \ref{prop:2} hold; 2) $\lim_{\epsilon\rightarrow 0}b(\epsilon)=0$; and 3) $b(0)=0$, then for any $\psi$, the limit of the gradients converge to zero as the discriminator $d_{\phi}$ converges to the optimal discriminator $d^{*}$: $\lim_{\Vert d-d^{*}\Vert\rightarrow 0} \nabla_{\psi}V=0$ and $\lim_{\Vert d-d^{*}\Vert\rightarrow 0} \nabla_{\psi}V_{emp}=0$.
	\end{corollary}
	\begin{remark}
		Examples of value functions in Corollary \ref{cor:1} are vanilla GAN with $b(t)=\log{(1-t)}$ and Wasserstein GAN with $b(t)=-t$. Note that AVO \cite{louppe2019adversarial} uses Wasserstein GAN in their released code.
	\end{remark}
	
	\subsection{Implicit Relation Problem of Proposal Distribution Approach}
	\label{sec:ImplicitRelation}
	
	The second problem is the implicit relation between the proposal distribution $p_{\psi}(\theta)$ and the posterior distribution $p(\theta|x_{obs})$. In the training time, the proposal distribution approximates $\mathbb{P}_{g}$ to $\mathbb{P}_{emp}$ instead of $\mathbb{P}_{r}$, at best. Proposition \ref{prop:3} analyzes marginal form when $\mathbb{P}_{g}$ equals to $\mathbb{P}_{emp}$.
	
	\begin{proposition}[\emph{Implicit Relation}]\label{prop:3}
		Assume that the generator distribution $\mathbb{P}_{g}$ equals to the empirical data distribution $\mathbb{P}_{emp}$. Then, the marginal equivalence holds between the input parameter distribution $p_{\psi}(\theta)$ and the posterior distribution $p(\theta|x_{obs})$: $\int p(x|\theta)p_{\psi}(\theta)d\theta = \int p(x|\theta)p(\theta|x_{obs})d\theta$.
	\end{proposition}
	Even though the proposal approach succeeds on estimating $\mathbb{P}_{emp}$ through $\mathbb{P}_{g}$, the marginal equivalence does not guarantee the equivalence between the optimal proposal distribution $p_{\psi^{*}}(\theta)$ and the posterior distribution $p(\theta|x_{obs})$. Figure \ref{img:Multi-modal} (d) illustrates an example of the non-equivalence between the implicit proposal $p_{\psi}(\theta)$ and the posterior $p(\theta|x_{obs})$ in AVO. This \textit{implicit relation problem} can be mitigated by estimating the likelihood $p(x_{obs}|\theta)$ directly to infer the posterior distribution $p(\theta|x_{obs})$.
	
	\section{Adversarial Likelihood-Free Inference} \label{sec:ALFI}
	
The \textit{implicit relation problem} causes AVO to be ill-posed, i.e. there could be many candidates for the optimal proposal distribution $p_{\psi^{*}}$ that achieves $\mathbb{P}_{g}$ to equal to $\mathbb{P}_{emp}$. Besides, the \textit{gradient vanishing problem} forces $p_{\psi}$ to remain to be pre-matured, so the convergence of $p_{\psi}$ to an optimal proposal $p_{\psi^{*}}$ is not guaranteed. Therefore, we introduce Adversarial Likelihood-Free Inference (ALFI), in place of proposal distribution approach, by breaking apart \textit{likelihood-free inference} into the likelihood estimation problem and the sampling problem. ALFI estimates the likelihood by Theorem \ref{thm:1} with a surrogate model parametrized by $\psi$, and ALFI proposes the next inputs via a sampling algorithm.

	ALFI consists of the three components: a non-optimizable black-box generator $g:\mathbb{R}^{d}\times\mathbb{R}^{m}\rightarrow\mathbb{R}^{q}$, a discriminator network $d_{\phi}:\mathbb{R}^{q}\rightarrow[0,1]$, and a probabilistic encoder network $s_{\psi}:\mathbb{R}^{d}\rightarrow\mathbb{R}^{K}$, where the probabilistic encoder network forms a surrogate model to estimate the likelihood. The generator $g(\theta,u|\omega)$ is a function of $\theta\in\Theta\subseteq\mathbb{R}^{d}$ and $u\in\mathbb{R}^{m}$, conditioned on the fixed generator internal coefficients $\omega$, throughout the inference stage, determined a-priori either by the domain experts or by another statistical model. Algorithm \ref{alg:ALFI} presents three procedures of ALFI: 1) sampling procedure in line 3; 2) evaluation procedure in line 4; and 3) learning prcedure in lines 6 and 7.

AVO suffers from \textit{gradient vanishing problem} because AVO formulates \textit{likelihood-free inference} as a saddle point problem with respect to $\psi$ and $\phi$ of an adversarial value function $V$. On the other hand, ALFI detours \textit{gradient vanishing problem} by formulating \textit{likelihood-free inference} as a pair of maximization problems of separated losses $\mathcal{L}_{d}(\phi)$ and $\mathcal{L}_{s}(\psi)$ on the discriminator $d_{\phi}$ and the encoder $s_{\psi}$, respectively. In particular, while the discriminator maximizes the adversarially designed GAN loss $\mathcal{L}_{d}(\phi)$, the encoder updates parameter $\psi$ via Maximum Likelihood Estimation of a loss $\mathcal{L}_{s}(\psi)$ for a surrogate model. Moreover, the introduction of probabilistic encoder $s_{\psi}$ in ALFI mitigates \textit{implicit relation problem} by estimating the likelihood directly.

ALFI chooses the Metropolis-Hastings algorithm as a sampling algorithm. The Metropolis-Hastings algorithm is a fast,  parallelizable and mathematically well-developed sampler, and we prove Theorem \ref{thm:convergence} that guarantees the convergence of inhomogeneous Markov chain \cite{levin2017markov} to the posterior $p(\theta|x_{obs})$.
	
	\subsection{Likelihood Estimation} \label{sec:LikelihoodApproximation}
	\begin{algorithm}[t]
		\KwRequire{Discriminator network $d_{\phi}$, Probabilistic encoder network $b_{\psi}$}
		\BlankLine
		\For{$t$ steps}
		{
			\For{$m$ steps}
			{
				Sample the next particles $\{\theta_{i}^{'}\}_{i=1}^{n}$ from the current particles $\{\theta_{i}\}_{i=1}^{n}$, using Metropolis-Hastings algorithm with acceptance ratio $\bar{A}(\theta',\theta)$ given by Eq. \ref{eq:acceptanceRatioInPractice}
			}
			Execute the black-box generator with each particle in $\{\theta_{i}^{'}\}_{i=1}^{n}$\\
			\For{$l$ steps}
			{
				Update discriminator network parameters $\phi$ from the gradient of Eq. \ref{eq:discriminator}\\
				Update probabilistic encoder network parameters $\psi$ from the gradient of Eq. \ref{eq:probabilisticEncoding}\\
			}
		}	
		\caption{Adversarial Likelihood-Free Inference (ALFI)}
		\label{alg:ALFI}
	\end{algorithm}
	
To estimate the intractable likelihood, we introduce Theorem \ref{thm:1}, which states that the likelihood is a density of a $1$-dimensional random variable $Y_{\theta}$, see Appendix B.
	\begin{theorem}\label{thm:1}
		The likelihood becomes $p(x_{obs}|\theta)=p_{Y_{\theta}}\big(d_{\phi}(x_{obs})\big)$, where $Y_{\theta}:\mathbb{R}^{m}\rightarrow \mathbb{R}$ is the random variable under the map $Y_{\theta}(u)=d_{\phi}\big(g(\theta,u|\omega)\big)$, if $\Vert d_{\phi}-d^{*}\Vert<0.5$ with the supremum norm $\Vert d\Vert=\sup_{x\in\mathbb{R}^{q}}\vert d(x)\vert$.
	\end{theorem}
	
	Since the support of the random variable $Y_{\theta}$ is restricted to the unit interval, i.e. $d_\phi(\cdot) \in [0,1]$, Corollary $\ref{cor:2}$ expands the support to any of either a bounded interval, a semi-infinite interval, or a whole real line, where Corollary \ref{cor:2} is an application of change of variables in the probability densities.
	\begin{corollary}\label{cor:2}
		Let $h:[0,1]\rightarrow\mathbb{R}$ be strictly monotonic and continuous with a non-zero derivative at $d_{\phi}(x_{obs})$, then the likelihood becomes $p(x_{obs}|\theta)=p_{Z_{\theta}}\big(h(d_{\phi}(x_{obs}))\big)\big\vert h'\big(d_{\phi}(x_{obs})\big)\big\vert$, where $Z_{\theta}=h(Y_{\theta})$ is a transformed random variable of $Y_{\theta}$ under $h$.
	\end{corollary}
	
	As the random variable $Y_{\theta}$ embeds the stochastic information of the stochastic nuisance variable $u$, $p_{Z_{\theta}}(z)dz$ is the probability of the random variable $h\big(d_{\phi}(g(\theta,u))\big)$ being observed in $[z,z+dz)$. Since the density on $Z_{\theta}$ is intractable because of the implicit nature of $u$, we use Corollary \ref{cor:3} to formulate the likelihood estimation problem as the shape parameter estimation problem by imposing an explicit parametric distribution on $Z_{\theta}$.
	
	\begin{corollary}\label{cor:3}
		If the random variable $Z_{\theta}$ follows a parametric probability distribution with shape parameters $\bm{s}_{\theta}=\{s_{\theta,k}\}_{k=1}^{K}$, the likelihood becomes
		\begin{align*}
		p(x_{obs}|\theta)=f\Big(h\big(d_{\phi}(x_{obs})\big);s_{\theta,1},...,s_{\theta,K}\Big)\big\vert h'\big(d_{\phi}(x_{obs})\big)\big\vert,
		\end{align*}
		where $f(\cdot;s_{\theta,1},...,s_{\theta,K})$ is the density of a parametric distribution with shape parameters $\{s_{\theta,k}\}_{k=1}^{K}$.
	\end{corollary}
	\begin{remark}
		An example of a parametric distribution is the beta distribution with shape parameters $\alpha$ and $\beta$, where $h$ is the identity function. The other example could be the Gaussian distribution with shape parameters $\mu$ and $\sigma$, where $h(y)=h_{0}\circ\tilde{h}(y)$ with $h_{0}(t)=\frac{-2\sin{2\pi t}}{1-\cos{2\pi t}}$ and $\tilde{h}(y)=1/(1+e^{-(y-d_{\phi}(x_{obs}))})$, or $h(y)=\tilde{h}^{-1}(y)$, see Appendix D for the detailed explanation on $h$.
	\end{remark}
	
	\subsection{Acceptance Ratio}\label{sec:acceptanceRatio}
	
	\begin{corollary}\label{cor:4}
		If the prior distribution on $\theta$ is uniform and the random variables $Z_{\theta}$ and $Z_{\theta'}$ follow a parametric probability distribution with shape parameters $\bm{s}_{\theta}=\{s_{\theta,k}\}_{k=1}^{K}$ and $\bm{s}_{\theta'}=\{s_{\theta',k}\}_{k=1}^{K}$, respectively, the acceptance ratio of the Metropolis-Hastings algorithm of jumping to $\theta'$ from $\theta$ is
		\begin{align}\label{eq:acceptanceRatio}
		A(\theta',\theta)=\min\bigg(1,\frac{f\big(h(d_{\phi}(x_{obs}));s_{\theta',1},...,s_{\theta',K}\big)}{f\big(h(d_{\phi}(x_{obs}));s_{\theta,1},...,s_{\theta,K}\big)}\bigg).
		\end{align}
	\end{corollary}
	If we define a stochastic process $\{Z_{\theta}|\theta\in\Theta\}$ to be the collection of the random variables $Z_{\theta}$, the \textit{optimal} shape parameters $\bm{s}_{\theta}$ become diverse by the random variables $Z_{\theta}$ with different parameters $\theta$. The probabilistic encoder network, $s_{\psi}(\theta)=\hat{\bm{s}}_{\theta}$, is a surrogate model that estimates the $K$-dimensional \textit{optimal} shape parameters $\bm{s}_{\theta}$ of the probability distribution for $Z_{\theta}$. 
	
	\subsection{Algorithm}
	
	At the $t$-th iteration, the Metropolis-Hastings algorithm samples the next $n$ independent set of parameters $\{\theta_{t+1}^{(i)}\}_{i=1}^{n}$. The $i$-th particle $\theta_{t}^{(i)}$ searches the neighborhood of $\theta_{t}^{(i)}$ by suggesting an intermediate particle $\tilde{\theta}_{t+1}^{(i)}$ from the symmetric proposal distribution. The intermediate particle $\tilde{\theta}_{t+1}^{(i)}$ will be accepted to be the next parameter $\theta_{t+1}^{(i)}$ by the below probability:
	\begin{align}\label{eq:acceptanceRatioInPractice}
	\bar{A}(\tilde{\theta}_{t+1}^{(i)},\theta_{t}^{(i)})=\min\bigg(1,\frac{f\big(h(d_{\phi}(x_{obs}));s_{\psi}(\tilde{\theta}_{t+1}^{(i)})\big)}{f\big(h(d_{\phi}(x_{obs}));s_{\psi}(\theta_{t}^{(i)})\big)}\bigg).
	\end{align}
	The ratio in Eq. \ref{eq:acceptanceRatioInPractice} equals to the true acceptance ratio in Eq. \ref{eq:acceptanceRatio} when the probabilistic encoder network $s_{\psi}$ estimates the exact shape parameters $\bm{s}_{\theta}$.
	
	After sampling the $n$ independent particles, ALFI puts each particle into the black-box generator to evaluate. Then, the discriminator classifies the $n$ generated fake data with the real data. To maximize $d_{\phi}(x)$ for $x\in\mathbb{P}_{r}$ and minimize $d_{\phi}(\tilde{x})$ for $\tilde{x}\in\mathbb{P}_{g}$, we use the Wasserstein loss \cite{arjovsky2017wasserstein}
	\begin{align}\label{eq:discriminator}
	\mathcal{L}_{d}(\phi)=-\mathbb{E}_{x\sim\mathbb{P}_{r}}\big[d_{\phi}(x)\big]+\mathbb{E}_{\tilde{x}\sim\mathbb{P}_{g}}\big[d_{\phi}(\tilde{x})\big].
	\end{align}
	We calculate the above expectation $\mathbb{E}_{\tilde{x}\sim\mathbb{P}_{g}}$ through the Monte-Carlo estimation with sampled fake data $\big\{g(\theta_{t+1}^{(i)},u_{t+1}^{(i)}|\omega)\big\}_{i=1}^{n}$, where $\{\theta_{t+1}^{(i)}\}_{i=1}^{n}$ are selected from the Metropolis-Hastings algorithm and $\{u_{t+1}^{(i)}\}_{i=1}^{n}$ are the sampled nuisance variables that are determined for each execution.
	
	The probabilistic encoder estimates the shape parameters by minimizing the negative log-likelihood of $h\big(d_{\phi}(g(\theta,u|\omega))\big)$ being observed from the parametric distribution with shape parameters $s_{\psi}(\theta)$. The below expectation $\mathbb{E}_{\theta,u}$ equals to the expectation $\mathbb{E}_{\tilde{x}\sim\mathbb{P}_{g}}$.
	\begin{align}\label{eq:probabilisticEncoding}
	\mathcal{L}_{s}(\psi)=-\mathbb{E}_{\theta,u}\Big[\log{f\Big(h\big(d_{\phi}(g(\theta,u))\big);s_{\psi}(\theta)\Big)}\Big].
	\end{align}
	
	\subsection{Convergence of Inhomogeneous Markov Chain}\label{sec:convergence}
	
	After $t$ iterations of learning, the sampling procedure from the Metropolis-Hastings algorithm is equivalent to sample from the transition kernel, $P_{t}(\theta'|\theta)=q(\theta'|\theta)\min\big\{1,\frac{p_{t}(\theta'|x_{obs})}{p_{t}(\theta|x_{obs})}\big\}+\delta(\theta'-\theta)\int q(\tilde{\theta}|\theta)\big(1-\min\big\{1,\frac{p_{t}(\tilde{\theta}|x_{obs})}{p_{t}(\theta|x_{obs})}\big\}\big)d\tilde{\theta}$. Here, $q(\theta'|\theta)$ is a symmetric proposal distribution of the Metropolis-Hastings algorithm and $p_{t}(\theta|x_{obs})$ is the approximate posterior at $t$-th iteration, where the approximate likelihood, $p_{t}(x_{obs}|\theta)=f\big(h(d_{\phi_{t}}(x_{obs}));s_{\psi_{t}}(\theta)\big)\big\vert h'\big(d_{\phi_{t}}(x_{obs})\big)\big\vert$, is estimated by Corollary \ref{cor:3}. The parameter update of the discriminator and the encoder networks causes the transition kernel to be adjusted for every iteration. Therefore, the standard theory on Markov chain with fixed transition kernel \cite{gamerman2006markov} is not applicable, which means that the Markov chain is no longer guaranteed to asymptotically follow the posterior distribution $p(\theta|x_{obs})$.

Once the Markov chain does not follow the posterior distribution, the parameter learning may not succeed to estimate the exact likelihood, since the area near the true parameter $\theta^{*}$ could not have been visited in the process of learning. Therefore, we provide a theoretic analysis on the limit behavior of inhomogeneous Markov chain with updating transition kernel, which guarantees the success on \textit{likelihood-free inference} through ALFI structure that integrates the Metropolis-Hastings algorithm with the likelihood estimation networks. Theorem \ref{thm:convergence} ensures the convergence of distribution for the Markov chain with trainable transition kernel $P_{t}$ to the posterior distribution, $p(\theta|x_{obs})$.
	\begin{theorem}\label{thm:convergence}
		Assume that $Z_{\theta}$ follows either beta or Gaussian distribution and the probabilistic encoder network asymptotically estimate the true shape parameters. With the minorization condition \cite{neklyudov2019implicit}, sufficiently large $m$ and continuously differentiable black-box generator with respect to $\theta$, the distribution of the inhomogeneous Markov chain uniformly converges to the posterior distribution
		\begin{align}
		\lim_{N\rightarrow\infty}\Vert P_{1}^{m}...P_{t}^{m}-P^{\infty}\Vert=0,
		\end{align}
		where $P$ is the transition kernel that has the posterior, $p(\theta|x_{obs})$, as the unique stationary distribution; and where the operator norm, $\Vert P_{1}^{m}...P_{t}^{m}-P^{\infty}\Vert=\sup_{\theta_{0}}\Vert \delta_{\theta_{0}} P_{1}^{m}...P_{t}^{m}-p(\cdot|x_{obs})\Vert_{TV}$, is the supremum of the total variation norm, $\Vert q\Vert_{TV}=\int \vert q(\theta)\vert d\theta$.
	\end{theorem}

	\begin{remark}
		See Appendix C for the general version of Theorem \ref{thm:convergence} and the proofs. The distribution $\delta_{\theta_{0}}P_{1}^{m}...P_{t}^{m}$ is the distribution of the inhomogeneous Markov chain after $t$ iterations of learning, starting at $\theta_{0}$. The uniform convergence of Theorem \ref{thm:convergence} states that the convergence speed of the Markov chain to the posterior distribution is uniform, i.e. the mixing time of the Markov chain to the posterior distribution $p(\theta|x_{obs})$ is uniform with respect to the initial point $\theta_{0}$.
	\end{remark}
	
	\section{Experiments} \label{sec:Experiments}
	
	\begin{table}[t]
		
		\caption{The performance of \textit{likelihood-free inference} algorithms. The boldface indicates the highest performance among algorithms. Rejection ABC takes simulation budget 10 times more than ALFI.}
		\label{tab:performance}
		\begin{center}
			\begin{small}
				\begin{sc}
					\begin{adjustbox}{width=\columnwidth,center}
						\begin{tabular}{lcccccccccc}
							\toprule
							& Tumor \cite{unni2019mathematical} & SIR \cite{diekmann2000mathematical} & Poisson \cite{evans2010partial} & Stokes \cite{temam2001navier} & NPA \cite{sterman2001system} & MA(2) \cite{chan2010autoregressive} & M/G/1 \cite{newell2013applications} & Wealth \cite{wilensky1998netlogo}\\
							\hline
							\multirow{2}{*}{\parbox{2.3cm}{Rejection ABC\\(Reference) \cite{fu1997estimating}}} & \multirow{2}{*}{\textbf{5.0}$\pm$1.3} & \multirow{2}{*}{4.0$\pm$1.4} & \multirow{2}{*}{1.6$\pm$0.4} & \multirow{2}{*}{1.3$\pm$0.2} & \multirow{2}{*}{1.9$\pm$0.7} & \multirow{2}{*}{2.3$\pm$1.0} & \multirow{2}{*}{2.4$\pm$0.5} & \multirow{2}{*}{\textbf{2.9}$\pm$0.6}\\
							&&&&&&&&\\\hline
							MCMC ABC \cite{marjoram2003markov} & 2.2$\pm$0.7 & 2.3$\pm$1.0 & 1.7$\pm$0.6 & 1.3$\pm$0.7 & 1.5$\pm$0.6 & 2.6$\pm$1.1 & 1.6$\pm$0.5 & 2.2$\pm$1.0\\
							SMC ABC \cite{sisson2007sequential} & 4.1$\pm$0.7 & \textbf{4.8}$\pm$2.5 & 1.6$\pm$0.8 & 1.3$\pm$0.5 & 2.1$\pm$1.1 & 2.4$\pm$0.6 & \textbf{3.2}$\pm$1.1 & 2.3$\pm$0.4\\
							BOLFI \cite{gutmann2016bayesian} & 0.9$\pm$0.7 & 1.4$\pm$1.0 & 0.5$\pm$0.3 & 0.9$\pm$0.4 & 0.7$\pm$0.6 & 1.1$\pm$0.9 & 1.0$\pm$0.6 & 1.1$\pm$0.9\\
							ROMC \cite{ikonomov2019robust} & 1.8$\pm$0.6 & 2.3$\pm$0.3 & 1.8$\pm$0.3 & 0.5$\pm$0.3 & 0.5$\pm$0.3 & 1.9$\pm$1.1 & 1.3$\pm$0.5 & 1.3$\pm$0.6\\
							AVO (Gaussian) \cite{louppe2019adversarial} & 0.7$\pm$0.6 & 0.8$\pm$0.5 & 0.1$\pm$0.5 & 0.1$\pm$0.5 & 0.3$\pm$0.4 & 0.7$\pm$0.5 & 0.3$\pm$0.4 & 0.6$\pm$0.4\\
							AVO (Implicit) \cite{louppe2019adversarial} & 1.7$\pm$0.8 & 1.6$\pm$0.8 & 1.1$\pm$0.5 & 0.4$\pm$0.2 & 1.2$\pm$0.4 & 1.5$\pm$0.5 & 1.0$\pm$0.1 & 0.9$\pm$0.3\\\hline
							ALFI-beta & 4.9$\pm$1.1 & 3.9$\pm$1.5 & \textbf{2.8}$\pm$1.0 & \textbf{2.2}$\pm$0.8 & \textbf{2.4}$\pm$0.5 & 3.0$\pm$0.5 & 2.7$\pm$0.7 & 2.5$\pm$1.4\\
							ALFI-Gaussian & 3.7$\pm$0.5 & 3.1$\pm$0.9 & 2.4$\pm$0.4 & 1.9$\pm$0.6 & 2.3$\pm$0.5 & \textbf{3.3}$\pm$0.9 & 2.6$\pm$0.4 & 2.0$\pm$0.8\\
							\bottomrule
						\end{tabular}
					\end{adjustbox}
				\end{sc}
			\end{small}
		\end{center}
		\vskip -0.1in
	\end{table}
	
	\begin{table}[t]
		\vskip -0.1in
		\caption{Computational complexity and the wall clock time of \textit{likelihood-free inference} algorithms, see Appendix F for the further discussion. Rejection ABC, MCMC ABC, SMC ABC and BOLFI present (simulation time/sampling time) of the Poisson simulation model for the wall clock time. Other algorithms present (simulation time/sampling time/optimization time).}
		\label{tab:timeComplexity}
		\begin{center}
			\begin{small}
				\begin{sc}
					\begin{adjustbox}{width=\columnwidth,center}
						\begin{threeparttable}
							\begin{tabular}{lccccccc}
								\toprule
								& Rejection ABC & MCMC ABC & SMC ABC & BOLFI & ROMC & AVO & ALFI\\\hline
								Sampling Complexity & O($d$) & O($d$) & O($dn^{2}$) & O($dt^{3}$) \cite{cheng2017variational,saputro2017limited} & O($dM^{3}$) & O($P$) & O($P$)\\
								Optimization Complexity & --- & --- & --- & --- & O($dLM^{2}$) & O($P+Q$) & O($P+Q$)\\\hline
								Wall Clock Time (Wealth)\tnote{*} & $29h/2s$ & $3h/15s$ & $3h/1h$ & $24m/14h$ & $25h/14h/18h$ & $20h/48s/9m$ & $3h/15s/78s$\\
								\bottomrule
							\end{tabular}
							\begin{tablenotes}
								\item[*] $h$: $hours$, $m$: $minutes$, $s$: $seconds$
							\end{tablenotes}
						\end{threeparttable}
					\end{adjustbox}
				\end{sc}
			\end{small}
		\end{center}
		\vskip -0.2in
	\end{table}
	
	\subsection{Simulations as Black-box Generative Models}
	
	\begin{figure*}[t]
		\centering
		\includegraphics[width=0.8\textwidth]{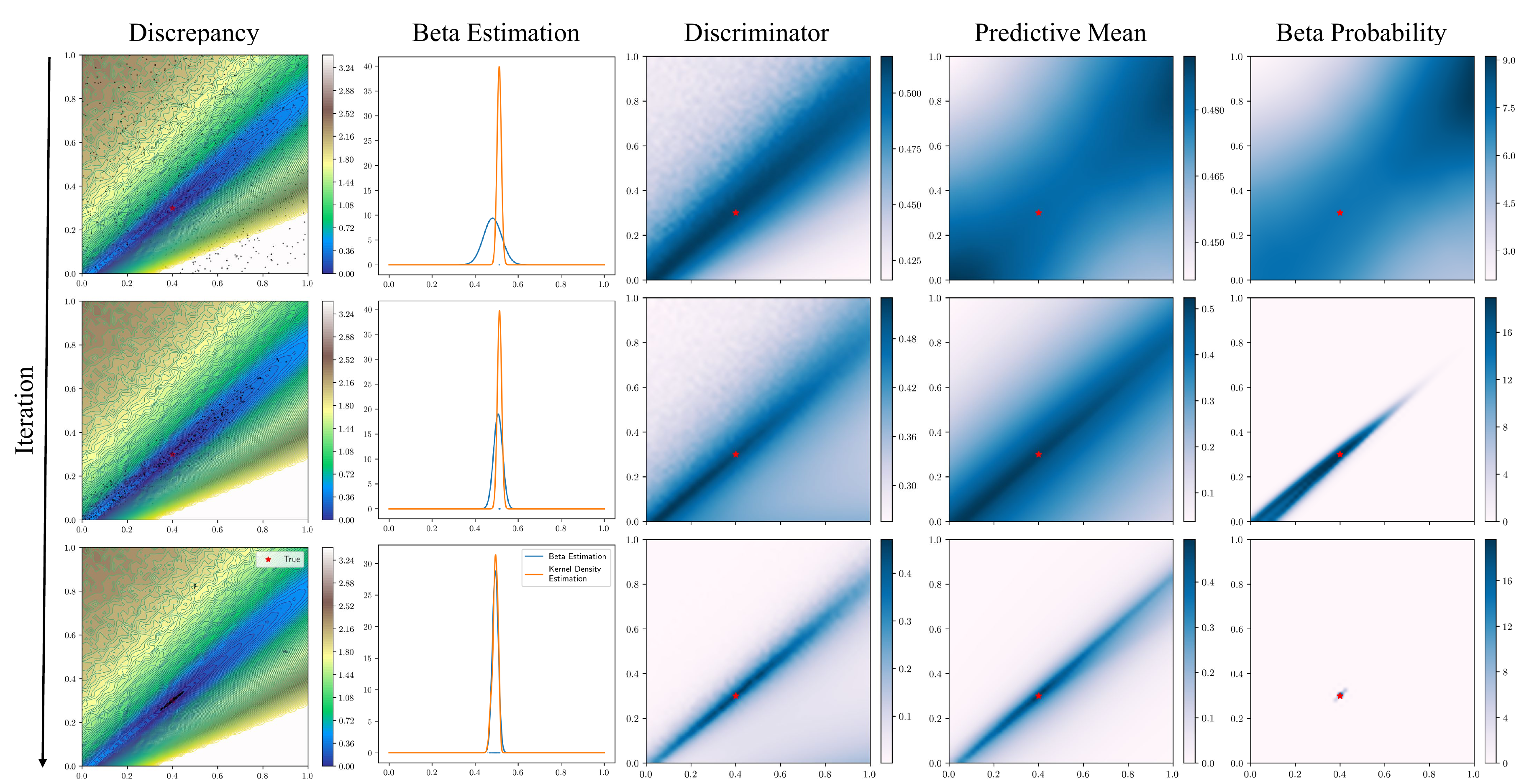}
		\caption{SIR model ALFI result (1st column) The Euclidean discrepancy landscape. The Metropolis-Hastings samples plotted as black dots (2nd column) The estimated beta distribution converges to the nonparametric kernel density estimation of $\big\{d_{\phi}(\theta^{*},u_{j})|j=1,...,100\big\}$ (3rd column) The discriminator vanish except $\{\theta|\Vert x_{obs}-g(\theta,u)\Vert_{2}<\epsilon\}$ for $\epsilon<<1$ (4th column) The mean of estimated beta distribution (5th column) The likelihood estimation concentrates to $\theta^{*}$}
		\label{img:Example2d}
		\vskip -0.15in
	\end{figure*}
	
	Simulation models with in-depth domain knowledge are the examples of the black-box generative models. Table \ref{tab:performance} presents the performance of \textit{likelihood-free inference} algorithms on eight simulation models (see Appendix E), where the performance is the negative log Euclidean distance, $\mathbb{E}_{\theta^{*}}\big[-\log{(\Vert\theta^{*}-\hat{\theta}\Vert_{2})}\big]$, between the true parameter $\theta^{*}$ and the estimated posterior mode $\hat{\theta}$. The observation $x_{obs}=\frac{1}{100}\sum_{j=1}^{100}g(\theta^{*},u_{j})$ is the average of 100 simulation executions. The algorithms are replicated for 10 times with a different set of true parameters $\{\theta_{k}^{*}\}_{k=1}^{10}$ to calculate the performance statistics. 
	
	\begin{wrapfigure}{r}{0.48\textwidth}
		\vskip -0.1in
		\begin{subfigure}{.32\linewidth}
			\centering
			\includegraphics[width=0.9\linewidth]{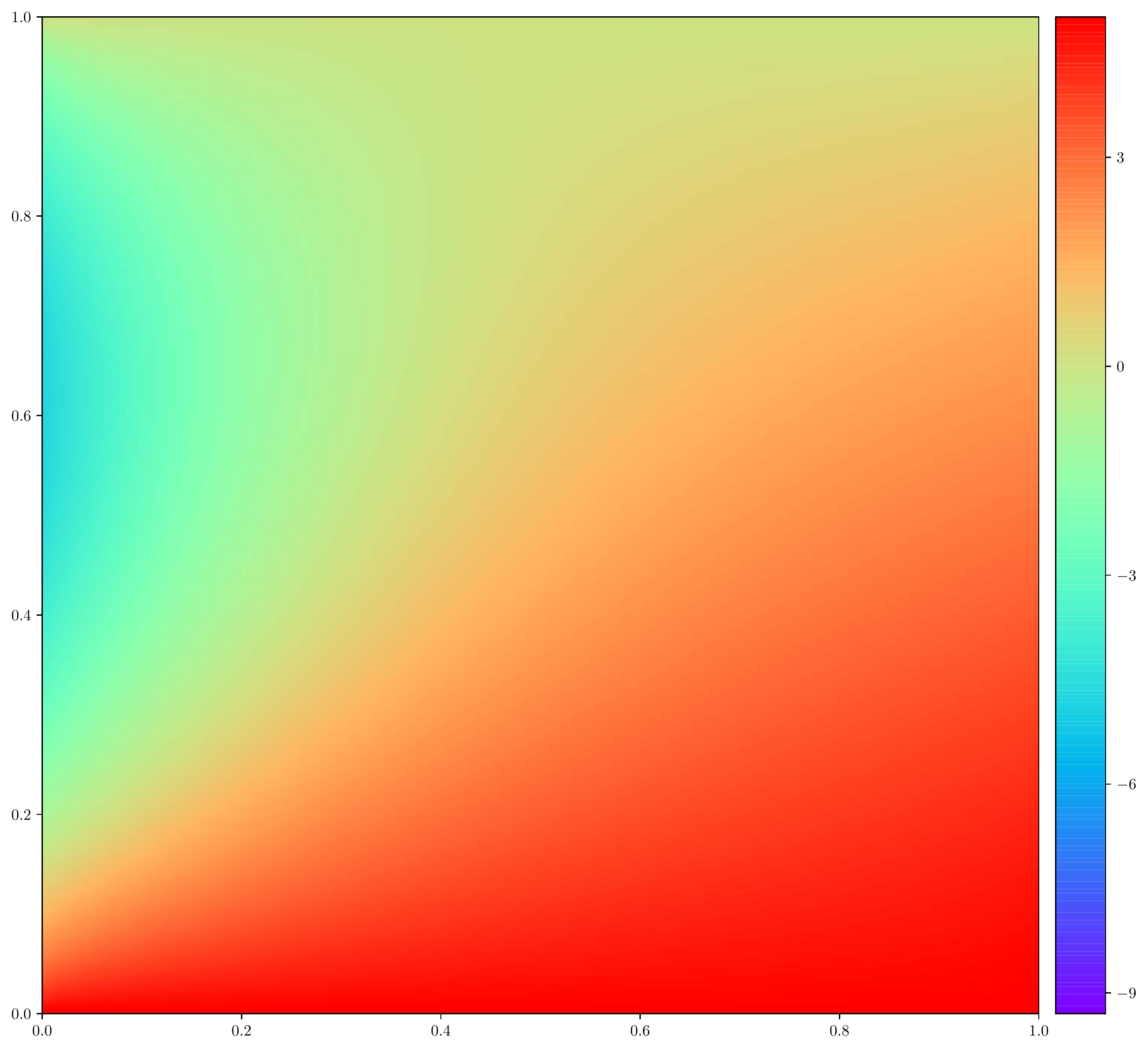}
			\subcaption{Observation}
		\end{subfigure}
		\begin{subfigure}{.32\linewidth}
			\centering
			\includegraphics[width=0.9\linewidth]{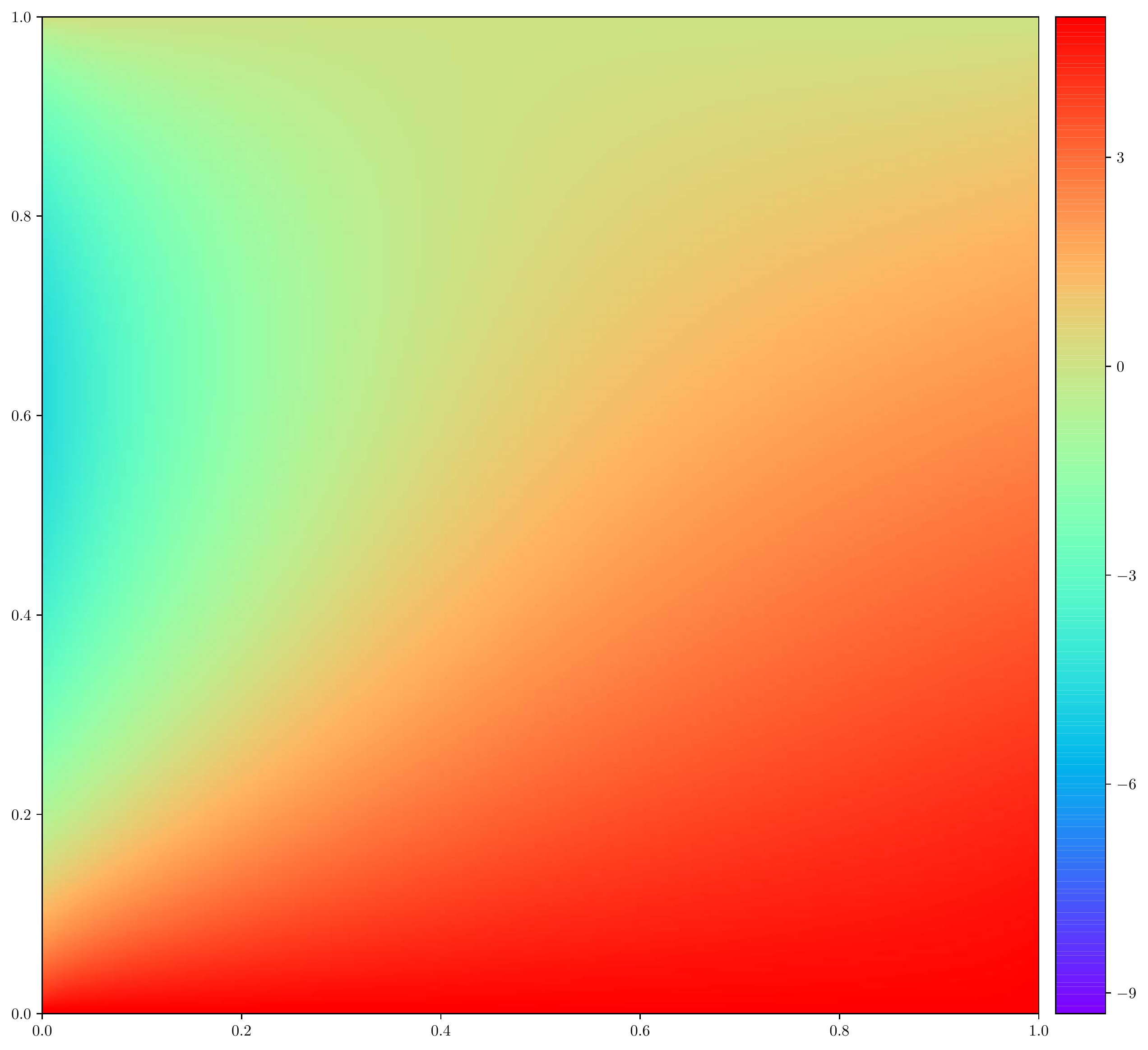}
			\subcaption{ALFI}
		\end{subfigure}
		\begin{subfigure}{.32\linewidth}
			\centering
			\includegraphics[width=0.9\linewidth]{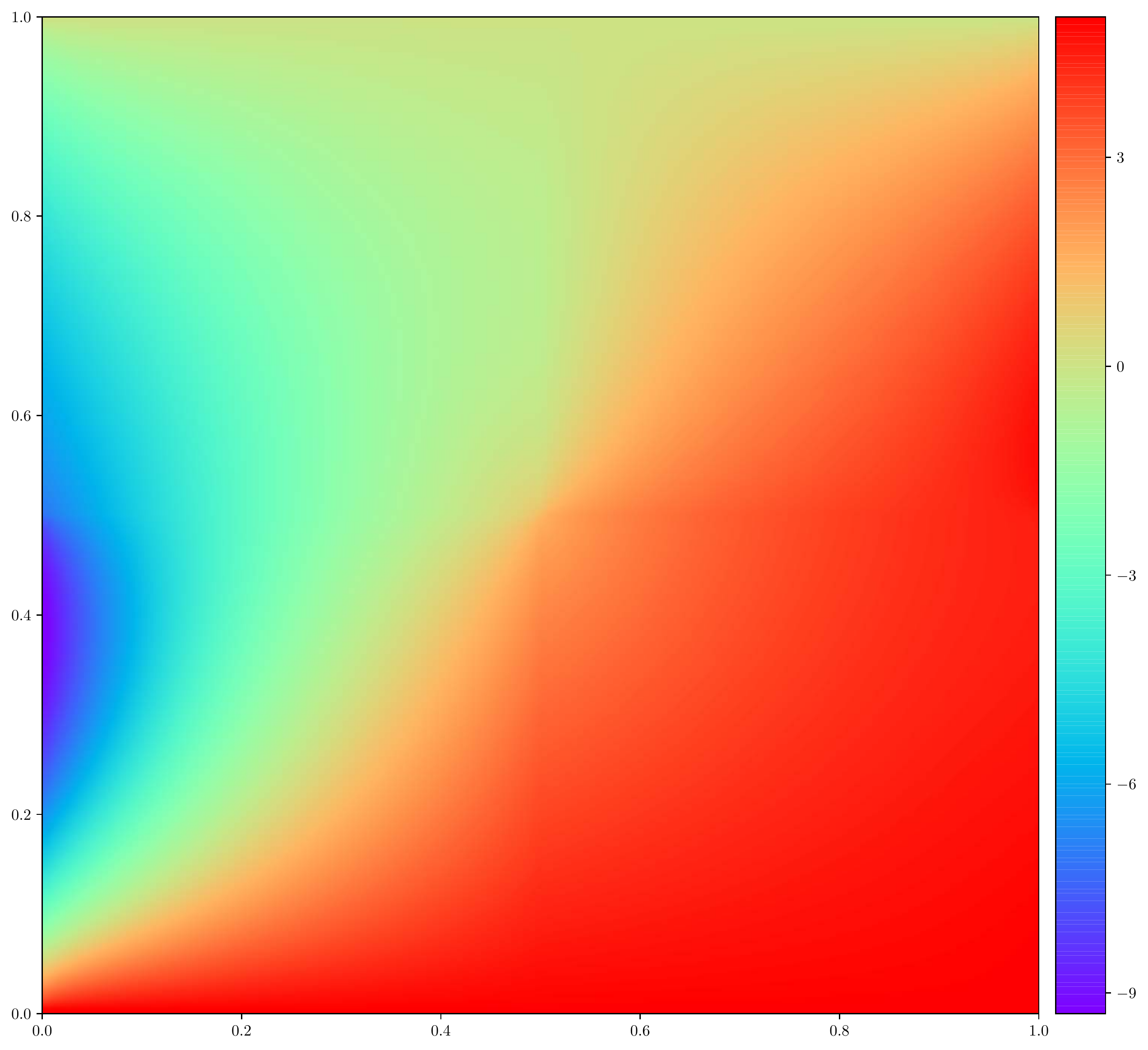}
			\subcaption{Implicit AVO}
		\end{subfigure}
		\caption{Simulation results}
		\label{img:Comparison}
		\vskip -0.1in
	\end{wrapfigure}
	
	Table \ref{tab:timeComplexity} presents the computational complexities; and the wall clock time of 1) simulation, 2) sampling and 3) optimization procedures for \textit{likelihood-free inference} algorithms. The Bayesian optimization-based algorithms, such as BOLFI and ROMC, take most of the computation time at the sampling procedure, whereas the sampling and the optimization time in ALFI are ignorable compared to the simulation time. Although ALFI allows parallelization by a cheap sampler, the sampling from parallel Bayesian optimization \cite{wu2016parallel} in BOLFI and ROMC is prohibitive due to the heavy computations in practice. Consequently, we take 100 simultaneous simulation executions in an iteration of ALFI, yet BOLFI takes a single simulation evaluation for an iteration. This property is important if a simulation is expensive.
	
	The second column of Figure \ref{img:Example2d} illustrates that the estimated beta distribution converges to the nonparametrically estimated density of $\big\{d_{\phi}(g(\theta^{*},u_{j}))|j=1,...,100\big\}$. The last column illustrates the contour map of the likelihood estimation, i.e. $Beta\big(d_{\phi}(x_{obs});s_{\psi}(\theta)\big)$, which concentrates to the true parameter $\theta^{*}$ after iterations. Figure \ref{img:Comparison} compares the Poisson simulation results among ALFI and implicit AVO; and Figure \ref{img:Comparison} concludes that ALFI is more identical to the observation than AVO.
	
	\subsection{Pre-trained Statistical Models as Black-box Generative Models}
	
	\subsubsection{Estimation of Spectral Density Mixture}
	
	\begin{wrapfigure}{r}{0.5\textwidth}
		\vskip -0.35in
		\begin{subfigure}{0.49\linewidth}
			\centering
			\includegraphics[width=\linewidth]{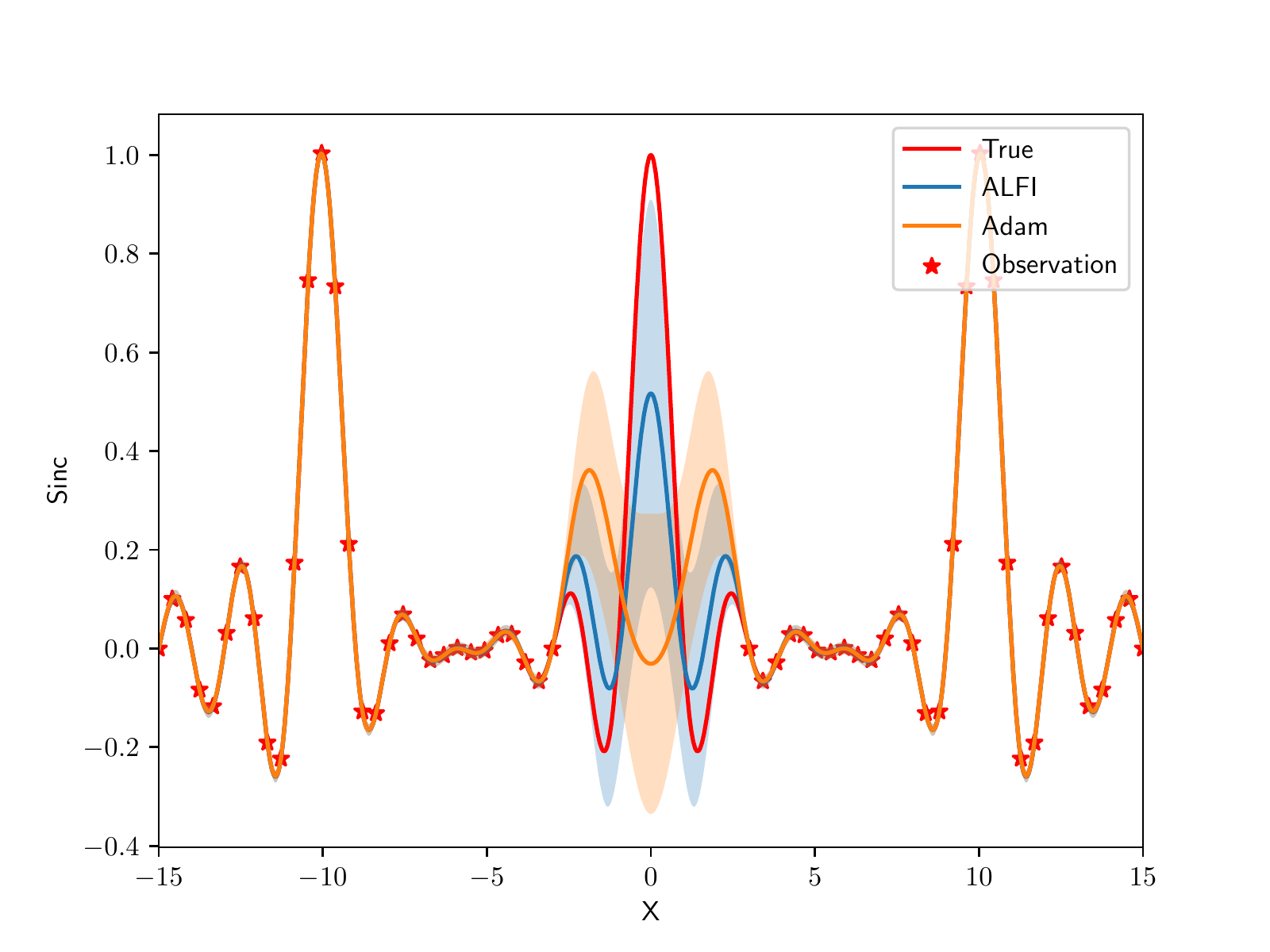}
			\subcaption{Interpolation}
		\end{subfigure}
		\begin{subfigure}{0.49\linewidth}
			\centering
			\includegraphics[width=\linewidth]{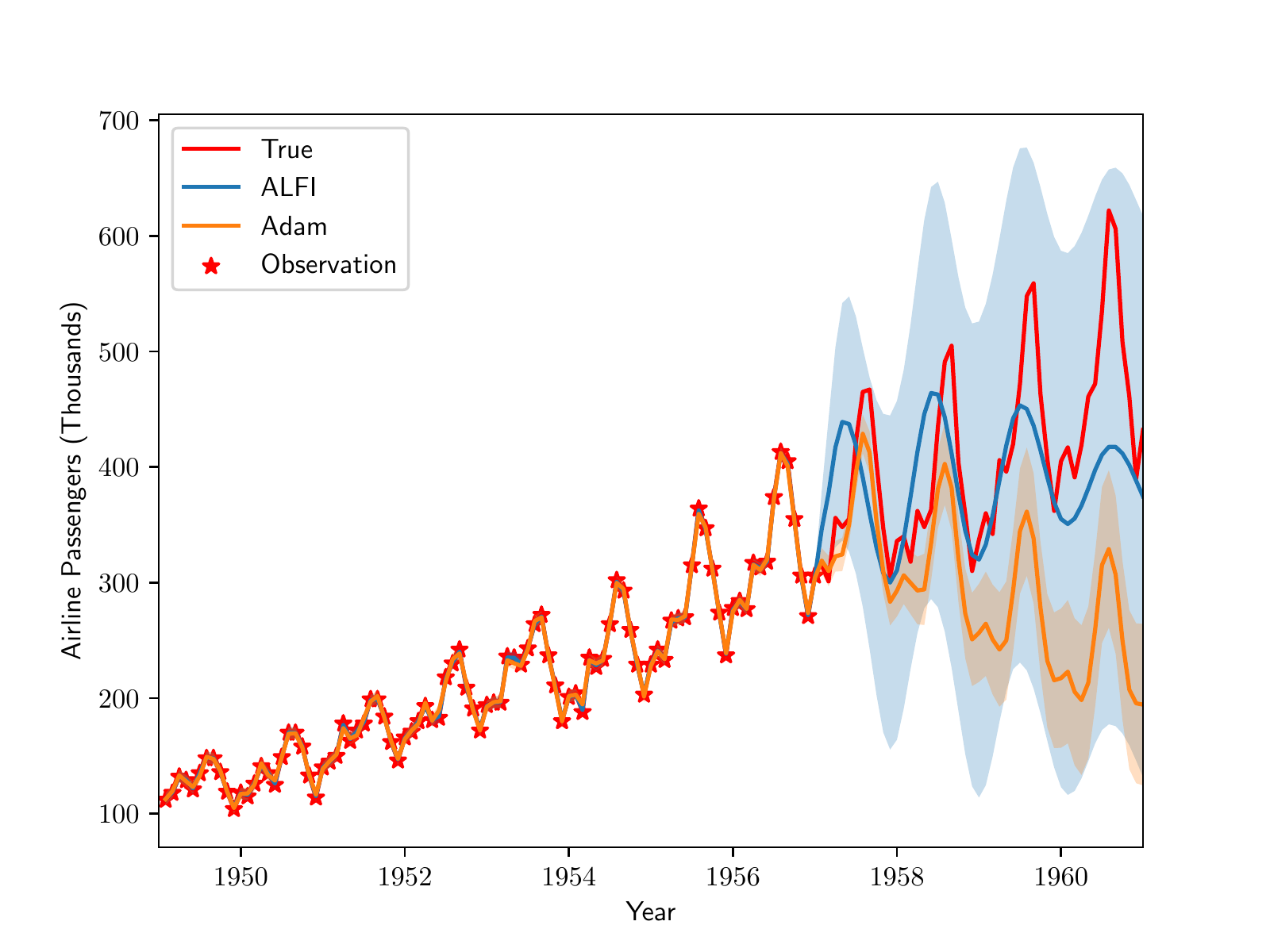}
			\subcaption{Extrapolation}
		\end{subfigure}
		\caption{ALFI interpolates/extrapolates better than Adam optimizer after 100 iterations of ALFI learning.}
		\label{img:SpectralMixtureEstimation}
		\vskip -0.2in
	\end{wrapfigure}
	
	We estimate the mixture parameters of the spectral density \cite{wilson2013gaussian} modeled by the mixture of Gaussian. The badly selected initial parameters make the optimization stuck at a local optimum with gradient descent. Considering ALFI as a gradient-free optimization algorithm, Figure \ref{img:SpectralMixtureEstimation} compares the interpolation/extrapolation results between the inference by ALFI and the gradient learning of the Adam optimizer. Figure \ref{img:SpectralMixtureEstimation} illustrates that ALFI optimizes the spectral mixture parameters better than Adam optimizer.
	
	\subsubsection{Corrupted Image Inpainting}
	
	\begin{wrapfigure}{r}{0.5\textwidth}
		\vskip -0.65in
		\begin{subfigure}{\linewidth}
			\centering
			\includegraphics[width=\linewidth]{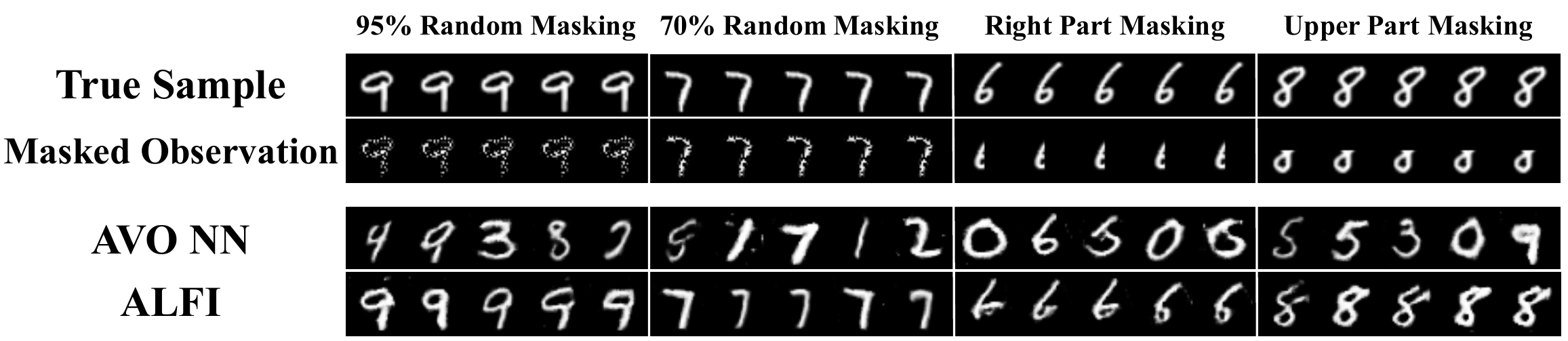}
			\subcaption{Masking}
		\end{subfigure}
		\begin{subfigure}{\linewidth}
			\centering
			\includegraphics[width=\linewidth]{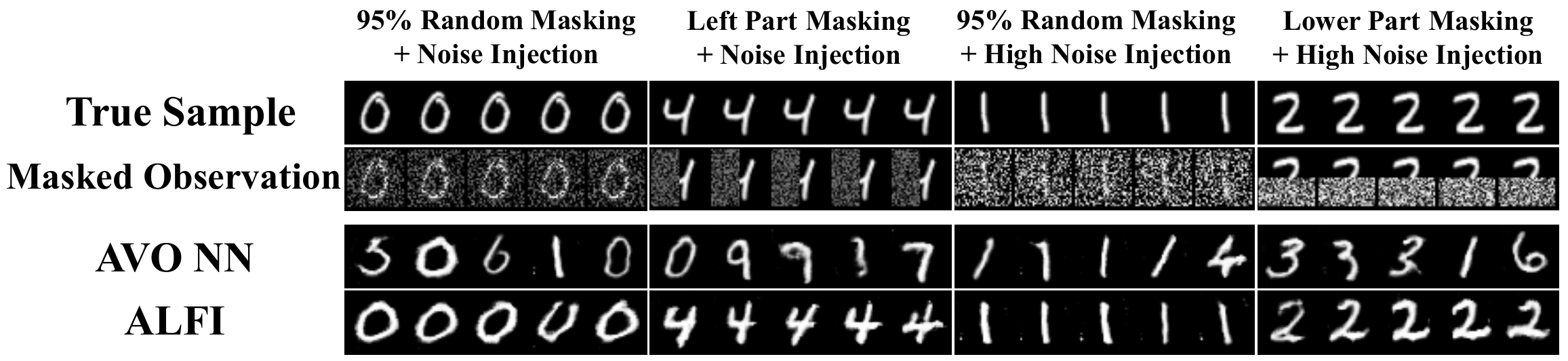}
			\subcaption{Noise Injection with Masking}
		\end{subfigure}
		\caption{ALFI regenerates better than implicit AVO for (a) masked and (b) noise injected masked images.}
		\label{img:Inpainting}
		\vskip -0.5in
	\end{wrapfigure}
	
	This experiment assumes a pre-trained DCGAN \cite{radford2015unsupervised} generator ($\omega$ is pre-trained) as the black-box model, and this experiment finds the best regeneration of the observation $x_{obs}$ by estimating the nearest latent embedding of $x_{obs}$. The observation $x_{obs}$ is a masked MNIST image. Figure \ref{img:Inpainting} concludes that ALFI can generate an image more similar to the true image than implicit AVO can generate.

	\section{Conclusions} \label{sec:Conclusions}
	The contribution of this paper is four-fold. First, this paper analyzes the \textit{gradient vanishing problem} and the \textit{implicit relation problem} of the previous research on \textit{likelihood-free inference}. Second, this paper provides a formula of the intractable likelihood as a $1$-dimensional density of a random variable in Theorem \ref{thm:1}. Third, this paper suggests a new \textit{likelihood-free inference} that uses the adversarial framework. Fourth, this paper suggests Theorem \ref{thm:convergence} that proves the convergence of the distributions of Markov chain to the posterior distribution, where the transition kernel of inhomogeneous Markov chain in the Metropolis-Hastings algorithm is dynamically updated.

\section*{Broader Impact}
We believe that ALFI is particularly useful in calibrating a simulation model to be realistic, and such simulations provide foundations for policymaking by what-if simulations. In this aspect, ALFI could aid policymakers in the decision-making process, by optimizing simulation models more congruent to the real-world system of interest. Additionally, ALFI allows us to infer unknown quantities in diverse domains, such as the diffusivity of a porous material; and this ability provides the profound efficiency in scientific simulations. However, we emphasize that ALFI is just an approximation of the posterior distribution, so the calibration using ALFI should not be fully trusted. Once ALFI fails to calibrate a simulation model and a practitioner depends exclusively on the simulation model in making policy, one might suggest a policy that could lead to catastrophic results.

	\bibliography{reference}

\end{document}